%% file: lrec2026-example.tex
\documentclass[10pt, a4paper]{article}
\usepackage{url}                
\usepackage[pdftex]{graphicx}   
\usepackage{caption,tabularx,booktabs} 
\usepackage{csquotes}           
\usepackage{subcaption}         
\usepackage{amsmath}            
\usepackage[T1]{fontenc}        
\usepackage{color,soul}         
\usepackage{threeparttable}     
\usepackage{threeparttablex}    
\usepackage{multirow}           
\usepackage{adjustbox}          
\usepackage[most]{tcolorbox}
\newcommand{\narrowbotc}[1]{{\colorbox{lightgray}{\parbox[t][][t]{\textwidth}{#1}}}} 

\usepackage[final]{lrec2026} 
\usepackage{cleveref}

\title{The Moral Foundations Reddit Corpus}

\name{\parbox{\textwidth}{
Jackson Trager\textsuperscript{*}, 
Alireza S. Ziabari\textsuperscript{*},
Elnaz Rahmati,
Aida Mostafazadeh Davani, 
Preni Golazizian, 
Farzan Karimi-Malekabadi,
Ali Omrani, 
Zhihe Li,
Brendan Kennedy, 
Georgios Chochlakis,
Nils Karl Reimer,
Melissa Reyes,
Kelsey Cheng,
Mellow Wei, 
Christina Merrifield,
Arta Khosravi,
Evans Alvarez,
and Morteza Dehghani}
}

\address{ 
University of Southern California \\
Corresponding author: jptrager@usc.edu
}

\abstract{Moral framing and sentiment can affect a variety of online and offline behaviors, including donation, environmental action, political engagement, and protest. Various computational methods in Natural Language Processing (NLP) have been used to detect moral sentiment from textual data, but achieving strong performance in such subjective tasks requires large, hand-annotated datasets. Previous corpora annotated for moral sentiment have proven valuable, and have generated new insights both within NLP and across the social sciences, but have been limited to Twitter. To facilitate improving our understanding of the role of moral rhetoric, we present the Moral Foundations Reddit Corpus, a collection of 16,123 English Reddit comments that have been curated from 12 distinct subreddits, hand-annotated by at least three trained annotators for 8 categories of moral sentiment (i.e., Care, Proportionality, Equality, Purity, Authority, Loyalty, Thin Morality, Implicit/Explicit Morality) based on the updated Moral Foundations Theory (MFT) framework. We evaluate baselines using large language models (Llama3-8B, Ministral-8B) in zero-shot, few-shot, and PEFT (Parameter-Efficient Fine-Tuning) settings, comparing their performance to fine-tuned encoder-only models like BERT (Bidirectional Encoder Representations from Transformers). The results show that LLMs continue to lag behind fine-tuned encoders on this subjective task, underscoring the ongoing need for human-annotated moral corpora for AI alignment evaluation.
 \\ \newline  }

\begin{document}

\maketitleabstract

\section{Introduction}
Moral rhetoric and framing play a role in increasing polarization and divisions in our societies \citep{marietta2008my,dehghani2016purity, brady2020mad}, but also in a wide range of pro-social behaviors that can potentially bring people together \citep{voelkel2022changing, wolsko2017expanding, kidwell2013getting, moaz2020using}. In order to understand the relationship between hate, division, compassion, and unity in the digital age, we need to understand the dynamics of moral language online. In particular, capturing and investigating the moral sentiment of text can allow  for the study of how individuals' and groups' expressed moral sentiment relate to various downstream online and offline behaviors.

Moral sentiment assessment and classification are subjective tasks, and when done automatically using Natural Language Processing (NLP) techniques, this subjectivity results in the need for large and diverse, both in terms of topics and coders, sets of annotations. The Moral Foundations Twitter Corpus \citep[MFTC; ][]{hoover2020moral}, a collection of 35,108 tweets that have been curated from seven distinct domains of discourse and hand annotated for 10 categories of moral sentiment (care, harm, fairness, inequality, loyalty, betrayal, authority, subversion, purity, and degradation) based on the Moral Foundations Theory \citep{haidt2012righteous,graham2011mapping}, was released a few years ago. This corpus has been used to design novel methods for moral sentiment classification \citep{asprino2022uncovering, lan2022text, burton2022understanding, araque2020moralstrength, wang2021different, wu2022unsupervised}, utilized in models to investigate the impacts of moral framing in other domains \citep[e.g., misinformation, polarization, and hate;][]{mutlu2020quantifying,abdurahman2025targeting,ruch2022millions, kennedy2023moral}, and has been applied to train models that produce morally salient text \citep[e.g., arguments and jokes;][]{alshomary2022moral, yamane2021humor}. In the era of large language models (LLMs), MFTC has been employed in evaluating the moral reasoning of LLMs \citep{ramezani2024moral} and serves as a foundation for new moral-reasoning benchmarks \citep{trager2025mftcxplain}, underscoring how morally annotated corpora support both evaluation and alignment work. However, as useful as MFTC is, its training dataset is limited to the social-media platform Twitter. 

Different online social media platforms have different linguistic and social structural environments that may result in variations in moral language and behavior \citep{curiskis2020evaluation}. Different platforms have varying character limits (e.g. 280 characters on Twitter compared to 10k-40k on Reddit) which alters the language usage of users \citep{boot2019character} and therefore may contribute to differences in the use and effectiveness of moral rhetoric \citep{Candia2022overuse}. Additionally, different platforms have different policies with respect to the levels of user anonymity and sensitive content moderation which may influence the domains of morality discussed given the potential judgements from others. Research has shown that higher levels of anonymity reduces the feeling of responsibility and alters moral behavior online \citep{simfors2020does}.
Lastly, modern NLP methods are known to require massive training data for producing sufficiently accurate, generalizable, and robust models. It has empirically been shown that diverse sets of training data, from different platforms and on different topics, can help improve the classification results by allowing the models to obtain generalized domain knowledge \citep[see][]{kennedy2022text}, rather than surface knowledge restricted to a particular platform and a small set of topics.

As mentioned previously, the MFTC relied on the Moral Foundation Theory's framework which is a pluralistic perspective of moral cognition and identifies multiple dimensions of moral values that have evolved to facilitate individual well-being, coalitional unity, and cooperation with strangers \citep{haidt2012righteous,graham2013moral}. The original version of the theory identified five separate but interrelated categories of moral concerns: Care/Harm, Fairness/Cheating, Loyalty/Betrayal, Authority/Subversion, and Purity/Degradation. Recently, a revision to the Moral Foundations Theory \citep{atari2023morality} split Fairness into the distinct and new foundations of Equality and Proportionality, while retaining the other four existing foundations. This split aims to capture the distinct moral concerns of fairness in procedure (proportionality) and equality of outcome (equality). In order to better understand the different nuances that will result from this theoretical change, we need to have an updated annotated corpus that complies with the latest theoretical revisions.

Together, the above reasons call for a corpus from a different platform, focused on a diverse set of topics, and annotated by a diverse group of annotators. Here we address this need by introducing the Moral Foundations Reddit Corpus (MFRC), a collection of 16,123 \url{Reddit.com} comments annotated for 8 categories of moral sentiment and curated from 12 morally-relevant subreddits.

Reddit is a public social media platform with over 52 million daily users who post content in over 138,000 active ``subreddits'' (user-created and user-moderated communities about different subjects) \citep{Reddit2021}. Shared content and comments in subreddits are ``voted'' on by users which is used to decide the visibility of the post. Activity on Reddit has been the center of numerous prominent cultural moments including coordinated attempts to challenge short-sellers of GameStop stock \citep{roose_2021} or attempts to identify the Boston bombing terrorists \citep{starbird2014rumors}. 

We focused our corpus compilation effort on Reddit for a number of reasons. First, in comparison to Twitter, Reddit shares many of the same research friendly features (e.g., responsiveness to current events, public posts, and available APIs) \citep{proferes2021studying}, but is organized into what are called subreddits. Different subreddits have distinct topics and consistent communities  \citep{datta2019extracting, soliman2019characterization} with varying cultures and norms \citep{chandrasekharan2018internet, fiesler2019ethical}. In relation to morality, these distinct communities and norms have led researchers to use Reddit to investigate moral conflicts across groups \citep[e.g.,][]{kumar2018community}, a phenomena that is harder to investigate on Twitter (which does not have organized groups) or Facebook (where many groups are private). Second, Reddit  provides more anonymity than many other social media platforms, potentially enabling users to more freely speak their minds and express their opinions \citep[e.g., ][]{triggs2021context,de2014mental, simfors2020does}. Third, in addition to general differences in language usage \citep{boot2019character}, the lack of restriction on the length of posts on Reddit may be particularly beneficial for training models. Fourth, we believe that Reddit has played a distinct role in contemporary politics. For example, the r/TheDonald and r/incels subreddits have been linked to political extremism \citep{gaudette2021upvoting} and mass shootings \citep{helm2022examining}.

To examine the effectiveness of LLMs for moral concern detection, we evaluate Ministral \citep{mistral2024ministral8b} and Llama \citep{dubey2024llama} under both zero-shot and few-shot prompting settings for single-label and multi-label text classification, following standard prompting methodologies \citep{NEURIPS2020_1457c0d6}. Beyond prompting, we also perform LoRA parameter-efficient fine-tuning \citep[PEFT;][]{hu2022lowrank} of Llama models for single-label and multi-label classification tasks \citep{tsoumakas2007multi}. For comparison, we include fully fine-tuned BERT models \citep{devlin2019bert}, trained in both single-label and multi-label configurations. These baselines allow us 
to understand the validity and relative performance of different text classification methods for identifying the moral concerns manifested in the MFRC.

Finally, in order to facilitate research into annotator response patterns and bias, as recommended by \citet{prabhakaran2021releasing}, we provide psychological and demographic metadata of our annotators. The background and biases of human annotators have been shown to impact their annotations \citep{hovy2021five,davani2022dealing,davani2023hate, bolukbasi2016man} with particularly damaging effects that amplify pre-existing biases \citep{mujtaba2019ethical, zhao2017men}. 
Annotators' biases may be particularly relevant in domains characterized by high subjectivity, such as moral values \citep{garten2019incorporating, garten2019measuring, hoover2020moral}. While, for example, an annotator’s political ideology might not have a substantial influence on how they annotate ``positive'' and ``negative'' sentiment in a corpus of hotel reviews, it seems likely that their ideology could substantially influence how they annotate expressions of justice and respect in a politically relevant corpora.


The contributions of this paper are threefold: (1) We introduce MFRC, a collection of 16,123 comments curated from 12 morally relevant subreddits and annotated for 8 moral sentiment categories based on the revised MFT. (2) We provide demographic and psychological metadata of annotators to facilitate future research on subjectivity and bias in moral sentiment annotation. (3) We evaluate a range of baseline models, including fine-tuned encoder-only models and recent LLMs in zero-shot, few-shot, and PEFT settings, establishing the MFRC as a benchmark for both moral sentiment classification and AI alignment evaluation. These contributions collectively extend prior work such as the MFTC \citep{hoover2020moral} to a new platform and theoretical framework, enabling richer cross-platform and cross-cultural analyses of moral language online.

\section{Corpus Overview}

As noted above, the MFRC consists of 16,123 Reddit comments drawn from 12 different subreddits. These subreddits were chosen based on the following criteria: First, we focused on subreddits that we expected to contain a wide range of moral concerns. Second, the chosen subreddits had to be active and have sufficient data. Third, we aimed to have a non-US based political subreddit with focus on current events that could be of use for different research communities. Accordingly, our corpus consists of 12 subreddits organized into three buckets; US politics, French politics, and Everyday moral life. The US politics bucket contains comments from 3 subreddits from the dates 1/1/2020 - 1/31/2021; \emph{r/politics} which captures political moral language generally, \emph{r/conservative} which covers moral rhetoric of the right, and \emph{r/antiwork} which covers different, but still political moral language from the left. The everyday moral life bucket is a collection of topics related to various aspects of everyday life, collected for their non-political moral judgement and moral emotions which includes comments from the 4 subreddits of \emph{r/nostalgia}, \emph{r/AmItheAsshole}, \emph{r/confession}, and \emph{r/relationshipadvice} between the dates of 1/1/2020 - 1/31/2021. The third bucket on French politics and contains comments from the  subreddits of \emph{r/europe}, \emph{r/worldnews}, \emph{r/neoliberal}, \emph{r/geopolitics}, and \emph{r/Conservative} that had the relevant keywords related to the presidential race including `Macron', `Le Pen', `France', `French', and `Hollande' (see below for the full set of keywords used) from the dates of 01/01/2017 to 06/30/2017 and had at least 10 likes/comments in order to control for sufficient engagement.

The MFRC is available for download as a  HuggingFace dataset \footnote{\href{https://huggingface.co/datasets/USC-MOLA-Lab/MFRC}{https://huggingface.co/datasets/USC-MOLA-Lab/MFRC}}.

\subsection{General Sampling Procedure}
In assembling the MFRC, we sampled Reddit posts from a larger set of each subreddit. Our initial filtering criteria selected posts of sufficient length (at least 10 tokens), and  de-selected any posts that were automatically marked by reddit as a bot, with the text ``I am a bot'' appended to the end of the post. The French politics bucket was also filtered for comments that mentioned at least one of the French presidential candidates (`macron', `le pen', `hollande', `dupont-aignan', `hamon', `arthaud', `poutou', `cheminade', `lassalle', `melenchon', `asselineau', `fillon') in which most hits were for frontrunners Macron, Le Pen, Hollande, and Fillon. For the US Politics and Everyday buckets, comments were selected that had a comment score of at least 10. 

In this filtered set, the proportion of comments that contained moral sentiment proved too low to conduct fully randomized sampling in a way that would result in a sufficient amount of moral examples. While these subreddits were chosen specifically for their potential moral salience, research has shown that use of moral language in some domains are rare \citep{atari2022}. To address this issue, a semi-supervised method was used to up-sample from moralized posts  \citep{kennedy2022introducing}.
Specifically, we first used word embeddings and a list of moral foundations seed words to compute a moral loading score \citep[Distributed Dictionary Representations;][]{garten2018dictionaries} for every comment. Next, for each moral concern, we computed the 95\% percentile scores to mark the highly moral comments and set bin size. 
Finally, in order to have a diverse range of moral posts, we compiled comments in a manner that each  subreddit bucket consisted of 1/2 comments with high moral loading ($> 95\%$), and 1/2 comments with less high moral loading ($\le 95\%$). 

This filtering and sampling procedure yielded approximately 6,000 comments for US Politics, 6,000 for Everyday Politics, and 8,000 comments for French politics. However, since vice, virtue, and multiple foundations regularly co-occur in an individual comment, duplicates occurred and were subsequently removed, resulting in a smaller final sample size for each bucket (US Politics: 4,821; French Politics: 6,489; Everyday Morality: 4,813; Total: 16,123).

\section{Annotation}
\label{sec:annotation}

Every post in the MFRC has been labeled by at least three trained annotators from a set of five (see Table~\ref{table:1} for the distribution of annotators for the corpus) for 8 categories of moral sentiment as outlined in the new version of our annotation manual (See Appendix \ref{sec:CodingGuide}).

\subsection{Moral sentiment}

Moral sentiment labels are drawn from the recently revised typology of Moral Foundations Theory \citep{atari2023morality}, which proposes a six-factor taxonomy of morality. In this model, each factor includes both virtues, or prescriptive moral concerns, and vices, prohibitive moral concerns. The proposed moral foundations are \citep[][]{atari2023morality}:

\textbf{Care/Harm:} Intuitions about avoiding emotional and physical damage or harm to others. It underlies virtues of kindness, gentleness, and nurturing, and vices of meanness, violence, and abuse.  

\textbf{Equality/Inequality:} Intuitions about egalitarian treatment and equal outcome for all individuals and groups. It underlies virtues of social justice and equality, and vices of discrimination and prejudice.  

\textbf{Proportionality/Disproportionately:} Intuitions about individuals getting rewarded in proportion to their merit (i.e., effort, talent, or input). It underlies virtues of meritocracy, productiveness, and deservingness, and vices of corruption and nepotism. 

\textbf{Loyalty/Betrayal:} Intuitions about in-group cooperation and out-group competition. It underlies virtues of patriotism and self-sacrifice for the group, and vices of abandonment, cheating, and treason.  

\textbf{Authority/Subversion:} Intuitions about deference toward legitimate authorities and high-status individuals. It underlies virtues of leadership and respect for tradition, and vices of disorderliness and resenting hierarchy. 

\textbf{Purity/Degradation:} Intuitions about avoiding bodily and spiritual contamination and degradation. It underlies virtues of sanctity, nobility, and cleanliness and vices of grossness, impurity, and sin.

Notably, unlike the MFTC, in the MFRC we did not code for Fairness. Rather, following the latest theoretical developments in the field \citep{atari2023morality}, we coded for separate foundations of Proportionality and Equality.

In addition to these six foundations, annotators were trained to look for an additional construct: \emph{Thin Morality} -- a moral judgment or concern which is voiced without clearly referring to one of the six moral domains \citep[][]{atari2022}. This brings the total categories of moral sentiment to 7. Annotators also had a formal category for \emph{Implicit/Explicit Morality } -- whether the moral sentiment in the comment was expressed explicitly or implicitly. Lastly, the annotators were asked to report  their overall level of confidence in their annotation as not confident, somewhat confident, or very confident. 

{\renewcommand{\arraystretch}{1.0}
\begin{table}[htp]
\centering
\footnotesize
\setlength{\tabcolsep}{4pt}
\begin{threeparttable}
\caption{Number of Reddit Posts Annotated by N Annotators for Each Subreddit}
\label{table:1}
\begin{tabular}{l|ccc}
\toprule
                                & \multicolumn{3}{c}{N Annotator} \\  
Subreddit                       & 3         & 4         & 5       \\ \midrule
r/AmItheAsshole                 & 1009      & 330       & -       \\
r/Conservative(French politics) & 75        & 69        & -       \\
r/Conservative(US politics)     & 870       & 898       & 8       \\
r/antiwork                      & 885       & 882       & 4       \\
r/confession                    & 993       & 338       & -       \\
r/europe                        & 1338      & 1302      & 7       \\
r/geopolitics                   & 53        & 59        & 1       \\
r/neoliberal                    & 846       & 815       & 12      \\
r/nostalgia                     & 994       & 348       & -       \\
r/politics                      & 864       & 894       & 10      \\
r/relationship\_advice            & 1021      & 331       & 1       \\
r/worldnews                       & 1267      & 1288      & 9       \\ \bottomrule
\end{tabular}
\end{threeparttable}
\end{table}}

\subsection{Annotators}
We started with a larger pool of 27 annotators, all undergraduate research assistants who completed two months of training sessions to develop expert-level familiarity with MFT. Training consisted of lectures, discussions, readings, and practice annotations with inter-annotation agreement analysis. In early annotation stages, annotator disagreement was addressed through discussion and, if necessary, certain labels were modified. However, given the subjective nature of the task, in many cases, it is difficult to make the determination of whether or not a document expresses moral sentiment or which category of moral sentiment it expresses. 
While it is necessary to have consistent annotator training, a focus on maximizing annotator agreement risks artificially inflating agreement at the cost of suppressing the natural variability of moral sentiment \citep{hoover2020moral}. Accordingly, our annotators were trained to both strive for consistency, while also encouraged to avoid stereotypes that may increase agreement with other annotators but would lead them to ignore their own beliefs.
Out of these original set of annotators, we selected the top five performing annotators, based on both inter-coder reliability assessments and the commitment of the annotators to the project, to become primary annotators and complete the rest of the annotations in our corpus.

Our trained annotators were independently assigned to label each comment from a subset of comments sampled from a corpus associated with one of the 12 subreddits (see Table ~\ref{table:5}). The annotations were performed on Prodigy.\footnote{\url{https://prodi.gy/}}  Each post was assigned a label indicating the absence or presence of the six foundations, thin morality, explicit/implicit expression, and the confidence level of the annotator (MFT Coding Manual in Appendix \ref{sec:CodingGuide}). 


\subsection{Annotator Metadata}

We have also collected responses to a range of psychological and demographic measures from our annotators. While keeping their identities anonymous, we provide measures of each annotator’s gender, 
sexual orientation, age,  household income, 
first language, 
political ideology along a liberal-conservative scale, religious affiliation, 
and moral values via the Moral Foundation Questionnaire-2 \citep{atari2023morality}.
Basic analyses demonstrate that our annotators' political ideology and morality skews liberal while family income skews wealthier than the average American. 
Based on the recommendations of \citet{prabhakaran2021releasing}, in order to increase utility and transparency of this corpus, these measures are provided by request (for privacy concerns) and encourage research into their potential impact on annotations, and the subsequent biases in the machine learning models. 


\section{Annotation Results}
The annotation results can be seen in Table ~\ref{table:5}. Recall that each post was annotated for multiple labels by multiple annotators, and the frequencies reported in Table \ref{table:5} are calculated based on annotators' majority vote (i.e. posts receiving at least 50\% agreement for that label). For example, if a particular post is annotated as `proportionality' by at least 50\% of the annotators who coded that post, then the majority vote on `proportionality' for that post is positive. We acknowledge the uneven distribution of labels across the various subreddits, in addition to the low base-rates of annotated posts for some the moral concerns, especially for Purity, Loyalty and Proportionality concerns.

\begin{table*}[ht]
\centering
\footnotesize
\setlength{\tabcolsep}{2pt}
\caption{Frequency of Reddit posts per Foundation Calculated Based on Annotators’ Majority Vote.}
\label{table:5}
\begin{tabular}{lcccccccc}
\toprule
{Subreddit} & {Care} & {Equality} & {Proportionality} & {Loyalty} & {Authority} & {Purity} & {Thin Morality} & {Non-Moral} \\
\midrule
r/AmItheAsshole & 343 & 145 & 103 & 56 & 50 & 38 & 227 & 456 \\
r/Conservative(French politics) & 12 & 5 & 6 & 7 & 16 & 0 & 18 & 95 \\
r/Conservative(US politics) & 195 & 200 & 84 & 38 & 155 & 45 & 231 & 945 \\
r/antiwork & 304 & 132 & 202 & 46 & 86 & 29 & 186 & 930 \\
r/confession & 281 & 69 & 101 & 24 & 52 & 50 & 249 & 574 \\
r/europe & 105 & 180 & 117 & 107 & 174 & 18 & 324 & 1741 \\
r/geopolitics & 1 & 4 & 5 & 1 & 5 & 0 & 9 & 100 \\
r/neoliberal & 39 & 74 & 67 & 50 & 117 & 11 & 195 & 1210 \\
r/nostalgia & 20 & 11 & 15 & 5 & 8 & 9 & 207 & 1160 \\
r/politics & 148 & 119 & 72 & 61 & 161 & 34 & 256 & 1016 \\
r/relationship\_advice & 418 & 137 & 84 & 98 & 31 & 56 & 191 & 450 \\
r/worldnews & 166 & 249 & 127 & 105 & 192 & 33 & 321 & 1543 \\
All & 2032 & 1325 & 983 & 598 & 1047 & 323 & 2414 & 10220 \\
\bottomrule
\end{tabular}
\end{table*}

\begin{figure}[!ht]
 \caption{The heatmaps show Interannotator Agreement (PABAK and Kappa) scores for all subreddits and foundations. Higher agreement corresponds with darker colors in both heatmaps.}
   \label{fig:interrater_aggreement}
  \centering
  \begin{subfigure}{\columnwidth}
  \centering
    \includegraphics[width=\columnwidth]{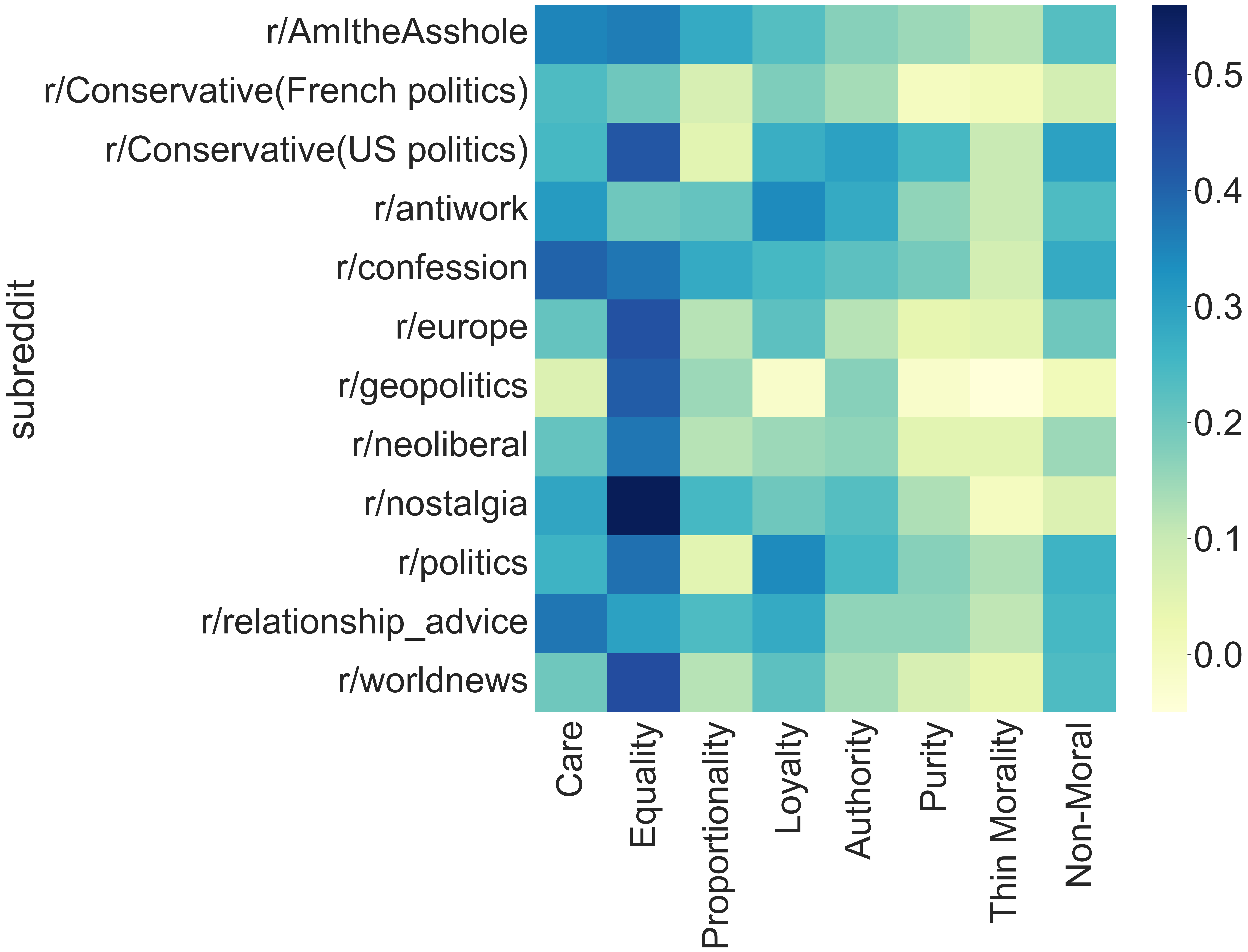}
    \subcaption{Fleiss’ kappa}
  \end{subfigure}
  \centering
  \begin{subfigure}{\columnwidth}
    \centering
    \includegraphics[width=\columnwidth]{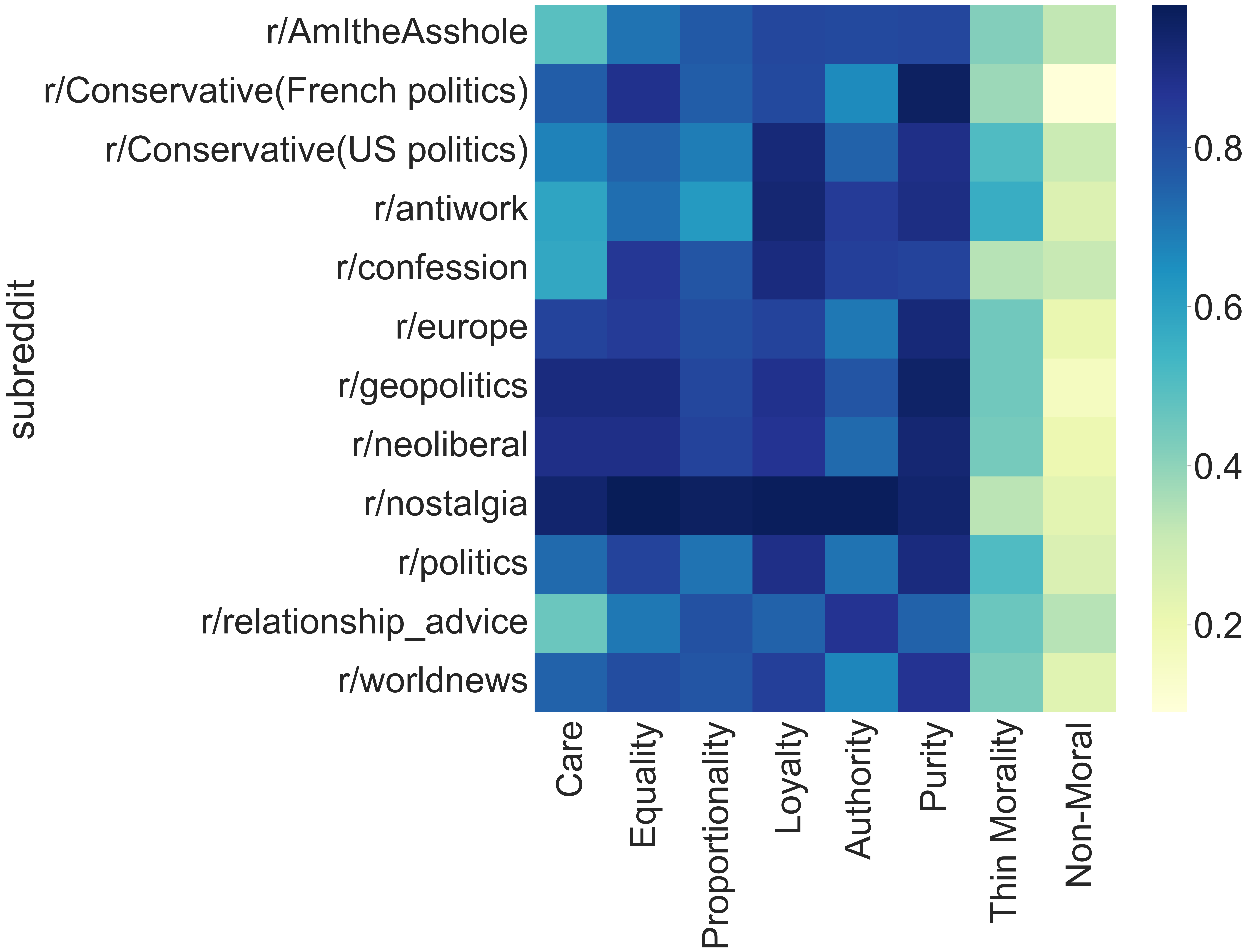}
    \subcaption{PABAK}
  \end{subfigure}
\end{figure}

The interannotator agreement results, using Fleiss’s \citep{fleiss1971measuring} kappa and prevalence-and bias-adjusted Fleiss’s kappa \citep[PABAK;][]{sim2005kappa} for multiple annotators, are displayed in Figure \ref{fig:interrater_aggreement}. 
Fleiss’s kappa is generally viewed as the gold standard measure for investigating agreement across many annotators, and it represents the degree of agreement beyond what is expected by chance. This measure though, is heavily influenced by the prevalence of positive cases. Given the subjective nature of our task, and the fact that positive cases are not prevalent given the often rarity of moral rhetoric \citep{atari2022}, we use PABAK which adjusts kappa for prevalence and bias. As expected, given the low base rate of moral language across the various subreddits (i.e. low positive cases), most reported kappa's are low. However, once adjusted for the issue of prevalence, we see medium to high agreements across the subreddits.


\section{MFRC as a Community Resource} 

The preprint release of MFRC has already been widely adopted across disciplines, serving as a shared empirical resource for research on morality, values, and language. Beyond being cited as related work, many studies have directly reused MFRC as training data, an evaluation benchmark, or a methodological reference point. These early applications provide evidence that MFRC fills a practical gap in the study of moral language and has begun to shape how researchers operationalize and evaluate moral constructs in computational settings.

A prominent line of work uses MFRC to evaluate and align LLMs with human moral judgments. For example, ~\citet{rathje2024gpt} include MFRC among multiple psychological datasets to benchmark GPT-3.5 and GPT-4 on moral foundation detection, comparing model outputs directly against human annotations. Similarly, ~\citet{abdurahman2024perils} analyze MFRC posts using both traditional NLP models and LLMs to demonstrate systematic divergences between machine-generated and human moral labels. More recent work explicitly leverages MFRC as a testbed for assessing the limits of LLM moral understanding, showing that even state-of-the-art models struggle with subtle or overlapping moral categories~\citep{bulla2025large,skorski2025beyond}. Collectively, these studies position MFRC as a grounding dataset for evaluating moral alignment and robustness in contemporary language models.

MFRC has also enabled methodological advances in moral NLP. Several studies use MFRC to train or benchmark new models for moral foundation classification, including domain-general classifiers such as MoralBERT~\citep{preniqi2024automatic}, multi-domain and transfer-learning approaches~\citep{guo2023data,chen2025mova}, and architectures enriched with emotional or event-level features~\citep{nguyen2024measuring,zangari2025me2}. Others exploit MFRC’s multi-label and subjective structure to study annotation disagreement and evaluation itself, proposing improved metrics or auditing techniques for noisy moral labels~\citep{mokhberian2022noise,chochlakis2025semantic}. Survey work further highlights MFRC as a central benchmark in the moral NLP literature, noting its consistent utility across platforms and domains~\citep{zangari2025survey}. Together, these efforts underscore MFRC’s role not only as a dataset, but as an infrastructure supporting model development, evaluation, and methodological reflection.

Finally, MFRC has served as a foundation for extending moral language analysis to new domains, languages, and modalities. Researchers have used MFRC-trained models to study moral disagreement in online discussions~\citep{van2023differences}, to probe the limits of English-centric moral classifiers in multilingual political texts~\citep{cheng2025beyond}, and to inspire domain-specific adaptations in areas such as news discourse, music lyrics, and multimodal content~\citep{preniqi2024automatic,lei2024emona}. Even when MFRC is not directly used, it frequently provides the moral taxonomy, baselines, or conceptual framing for new datasets and tasks. This breadth of uptake suggests that MFRC is already functioning as a general-purpose community resource, supporting cumulative research on how moral values are expressed, modeled, and evaluated in language.

\section{Baseline Classification Models}
In addition to compiling the corpus, we experiment with different models and provide baselines for predicting the annotators' majority vote of moral sentiment categories. Our goal is to simply establish baselines that future work can build on.

It should be noted that each post in MFRC was coded by multiple annotators for 8 different categories of moral sentiment discussed under the Section \ref{sec:annotation}. This is a multi-label classification task; i.e., not only are the categories of moral sentiment not independent of one another, but understanding variance in one domain should theoretically inform about  other related moral domains.  Here though, we provide both single-label (treating labels as independent) and multi-label classification results. 



\paragraph{Zero-shot and Few-shot Prompting}
To establish an initial baseline, we evaluate two widely used open-source LLMs, Meta-Llama-3-8B \citep{dubey2024llama} and Ministral-8B-Instruct-2410 \citep{mistral2024ministral8b}, using zero-shot prompting for both single-label and multi-label classification. We further explore few-shot prompting to assess whether including example instances for in-context learning \citep{NEURIPS2020_1457c0d6} improves performance. For this setup, we select five stratified samples based on label distribution and include them in the prompt. 

\paragraph{Classification Fine-tuning}
To examine the impact of training on LLM performance, we fine-tune the Llama model using parameter-efficient training \citep{hu2022lowrank} and evaluate its improvement on our dataset. Prior work has shown that for discriminative tasks, encoder-only models with full fine-tuning can outperform LLMs even after training \citep{qorib-etal-2024-decoder,roccabruna-etal-2024-will}. Therefore, we also include results from fine-tuning the BERT model \citep{devlin2019bert}. Since this model is smaller, we apply full fine-tuning (FFT) instead of PEFT. 

Due to the sparsity of our data, with far fewer moral posts than non-moral posts, we employ a weighted loss function. In this setup, the weight of sample $i$ for label $l$ is inversely proportional to the frequency of that label:

\begin{equation}
w_{i,l} \propto \frac{1}{\text{number of samples with label}~l}
\end{equation}





The baseline metrics we report are the F$_1$, precision, and recall, which are calculated across stratified 5-fold cross-validation. For the multi-label models, we stratified the data based on the moral sentiment label (the union of all the moral labels). 
We used 10\% of the training data for validation, and we chose the best performing model based on binary F$_1$ for single-label classification and F$_1$ macro for multi-label BERT. We train LLMs with a learning rate of 1e-5, batch size of 8, and 4 epochs, while BERT models are trained with a learning rate of 2e-5, batch size of 8, and 5 epochs. For multi-label classification, we double the number of training epochs. Learning rates are determined empirically from a search over the range [1e-4, 1e-6].

\section{Results}
The results of the baseline models are provided in Table \ref{tab:model_f1}. We provide both the average F1 score and standard deviations for each moral category (All metrics including recall and precision are included in the appendix \ref{sec:BaselineResults}). We ran each of the baseline models once for the entire corpus. Consistent with prior work \citep{qorib-etal-2024-decoder, roccabruna-etal-2024-will}, BERT models outperform the fine-tuned Llama models in terms of F$_1$.  Interestingly, the multi-label BERT model performed worse than its single-label counterpart, whereas the multi-label trained Llama achieved comparable or better results than the single-label version.

We then evaluated the best performing BERT models on the three subreddit buckets separately (French Politics, Everyday, and U.S. Politics). Consistent with previous results, BERT (single-label) outperformed BERT (multi-label) across all moral categories in all three buckets. However, performance varied across subreddits: certain moral foundations were predicted more accurately within the Everyday bucket, while others showed higher scores in U.S. Politics. This pattern suggests that subreddit focus and community context influence model performance, reflecting how moral language varies across online domains.

While our results establish majority-vote baselines, moral judgments remain inherently subjective, with meaningful annotator  heterogeneity. Motivated by concerns about aggregation in subjective tasks \citep{chochlakis2025aggregation}, we additionally examine annotator-level personalization. Personalization yields modest gains, particularly for annotators whose labeling patterns diverge more from the majority vote, but does not alter the overall pattern of findings, including the superiority of fine-tuned models. Details are provided in Section~\ref{sec:personalization} of the Appendix.

\input{tables/MFRC_f1}

\section{Cross-Corpus Classification}
\label{ssec:first}
In this section, we evaluate preliminary results on the transferability of models between MFTC and MFRC. Our hope is that future work can build on such cross-domain tasks in order to extract more generalized knowledge about moral rhetoric independent of the source and topic of the post.

In order to provide a fair comparison between MFRC and MFTC, we trained BERT models with the same hyperparameter discussed in the previous section. Additionally, we downsampled the MFTC dataset to have the same number of samples for each label as MFRC. The results are presented in Table \ref{table:MFTC}. In general, models trained and tested on the MFTC have better classification performance than the results for the MFRC (See Table \ref{tab:model_f1}).

\begin{table}[ht]
\centering
\footnotesize
\caption{BERT Results on MFTC}
\begin{tabular}{lccc} 
\toprule
Foundation  & F1 & Precision & Recall \\
\midrule
Authority & 0.65(0.02) & 0.65(0.04) & 0.65(0.05) \\
Care & 0.75(0.02) & 0.76(0.03) & 0.74(0.03) \\
Fairness & 0.82(0.01) & 0.83(0.04) & 0.81(0.03) \\
Loyalty & 0.58(0.05) & 0.66(0.09) & 0.52(0.06) \\
Purity & 0.54(0.04) & 0.63(0.05) & 0.48(0.05) \\
\bottomrule
\end{tabular}
\label{table:MFTC}
\end{table}

As mentioned before, the MFRC relies on the updated taxonomy of MFT in which the Fairness foundation is split into Proportionality and Equality. This makes the cross-corpus training for Fairness, Proportionality and Equality difficult. We hope the MFTC will be updated to use the more nuanced Proportionality and Equality labels in that corpus. For now though, for predicting Fairness labels in the MFTC, the union of  Proportionality and Equality labels are calculated based on the MFRC trained models, and the output of the union is then compared to the Fairness category in the MFTC. Similarly, to evaluate the MFTC Fairness models on the MFRC, both Proportionality and Equality are assigned the Fairness label predicted from the MFTC.  

\begin{table}[ht]
\centering
\small
\setlength{\tabcolsep}{10pt}
\caption{BERT Models Trained on MFRC, Evaluated on MFTC}
\begin{tabular}{lccc } 
\toprule
Foundation & F1 & Precision & Recall \\
\midrule
Authority &  0.38 & 0.53 & 0.30 \\
Care &  0.53 & 0.58 & 0.48 \\
Fairness &  0.35 & 0.76 & 0.23 \\
Loyalty &  0.38 & 0.50 & 0.31 \\
Purity &  0.28 & 0.56 & 0.18 \\
\bottomrule
\end{tabular}
\label{table:MFRConMFTC}
\end{table}

\begin{table}[ht]
\centering
\small
\setlength{\tabcolsep}{10pt}
\caption{BERT Models Trained on MFTC, Evaluated on MFRC}
\begin{tabular}{lccc}
\toprule
Foundation  & F1 & Precision & Recall \\
\midrule
Authority &   0.31 & 0.23 & 0.44 \\
Care &   0.43 & 0.53 & 0.35 \\
Fairness &   0.34 & 0.36 & 0.32 \\
Loyalty &   0.32 & 0.48 & 0.23 \\
Purity &   0.34 & 0.52 & 0.25 \\
\bottomrule
\end{tabular}
\label{table:MFTConMFRC}
\end{table}

Cross-corpus results are presented in Tables \ref{table:MFRConMFTC} and \ref{table:MFTConMFRC}. These preliminary results are indeed encouraging in that we demonstrate transferability between the two corpora in predicting out-of-domain distributions. Previous research training classifiers on the MFTC, and testing on Reddit \citep[e.g., ][]{atari2021morally} have shown similar levels of accuracies for cross-domain classification.
Any cross-corpus investigation should take into account the different time periods in which these two corpora were compiled. This difference can potentially impact the topics, the sentiments expressed about the topics, and the type of justification and reasoning used for the expressed sentiments. 
We believe though that more advanced methods in knowledge capture and representation could use the two corpora together to further achieve more generalized and better performing models.

\input{tables/New_table}

\section{Discussion}

Moral rhetoric and framing have been shown to be predictive of various important pro-social and anti-social behaviors. Several NLP methods have recently been proposed for capturing and categorizing moral sentiment based on textual data \citep[for a review see][]{atari2021chapter}. However, this is a subjective sentiment analysis task for which training data plays a vital role. To facilitate further research in this domain, here we introduced the MFRC, a collection of 16,123 Reddit comments annotated for 8 categories of moral sentiment, and provided a number of baseline results for different NLP models trained to predict moral sentiment. 

The MFTC was introduced in 2020, and so far this corpus has already facilitated multiple lines of research in both NLP and the social sciences. We believe Reddit's distinct linguistic and social structure, along with MFRC's methodological and theoretical updates, allow for potential new research that can both improve and expand the applications of MFTC. Specifically, the increased character limits on Reddit compared to Twitter is important for the more naturalistic expressions of moral rhetoric and its potential impact on the performance of classification models. While social media often provides large amount of data needed for training NLP models, with respect to sentiment analysis of moral language, the paucity of moral rhetoric in some domains  \citep{atari2022} makes it difficult to gather sufficient amounts of training data \citep{hoover2020moral}. Given Reddit's longer posts, models trained on the MFRC may perform better in out of domain tasks, especially in longer documents  (e.g., articles or speeches compared to tweets). Further, the distinct subreddit communities allow for the study of group linguistic dynamics. For example, shifts in moral language over time associated with a hashtag on Twitter may show the evolving general public opinion on a topic, while shifts in moral language within a particular subreddit may reflect the changing views of a specific community. 

Our results also showed that model performance varied across subreddits, suggesting that Reddit’s community-based structure represents a distinct linguistic and moral domain compared to other platforms. Recognizing these structural differences is important for interpreting the downstream group behaviors associated with moral language use on different platforms, such as voting and political mobilization. Moreover, the increased anonymity on Reddit can facilitate research into the gap between identity-linked and publicly-expressed moral concerns, such as on Facebook or Linkedin, and anonymous expressions of moral values, as on Reddit. 

As discussed previously, another important feature of the MFRC is that it is based on the newly updated version of the Moral Foundations Theory \citep{atari2023morality} which breaks the Fairness concern into the distinct moral concerns of Proportionality and Equality. The MFRC can be used to further investigate these nuances of fairness across topics such as income inequality and excessive wealth \citep{trager2025immorality} .   

Similar to the MFTC, the MFRC has detailed meta-data on the corpus annotators. We hope that by providing demographics and several key psychological measurements of our annotators, MFRC can facilitate future research into how annotators background characteristics impact their annotations.


As shown in Table~\ref{tab:model_f1}, fine-tuned models (BERT and PEFT-Llama) substantially outperform the zero-shot and few-shot baselines across all moral categories on MFRC. This pattern is consistent with the view that subjective, label-rich moral classification benefits most from domain-specific supervision, reinforcing the value of large, human-annotated corpora for both evaluation and downstream alignment work \citep{abdurahman2024perils}.

In conclusion, we hope that the MFRC, along with this report, in addition to the other previously released corpora in this domain \citep[e.g., ][]{kennedy2022introducing,hoover2020moral, trager2025mftcxplain}, can aid researchers by providing much-needed data and open new lines of research both in NLP and in the social sciences.  
In an age where political movements, grassroots activism, and plans for insurrections seem to take place in online environments, it is vital that we can better understand the moral dynamics of these online conversations. The intent of this project has been to further facilitate research into these timely topics.

\section{Data Disclaimer} We acknowledge that the compiled dataset contains biases and is not representative of diverse moral concerns present in world populations. Potential biases in the data include: biases specific to English-speaking countries and the English language, biases inherent to \url{Reddit.com} and its user base, biases in the researchers’ criteria for corpus curation as well as the underlying MFT itself, bias in the assessment of moral labels, and the fact that annotators were all undergraduate research assistants at a private academic institute. All of these factors, among others, likely influenced the annotations, as well as the performance of machine learning models trained on the corpus. Anyone using this corpus should be aware of these limitations and should acknowledge and/or try to mitigate them to the extent possible.

\section{Ethics Statement}
This work adheres to ethical standards for research involving human annotations and publicly available data. All annotators participated voluntarily, were compensated fairly, and received extensive training on respectful engagement with sensitive moral and political content. Annotation sessions included monitoring for emotional well-being and debriefing opportunities to mitigate potential distress. The study protocol was reviewed and approved by the authors’ institutional ethics board. We highlight that models trained on this corpus should be used only for scientific and educational purposes, and not for applications that could profile, surveil, or harm individuals or communities.\\

\section{Bibliographical References}\label{sec:reference}

\bibliographystyle{lrec2026-natbib}
\bibliography{lrec2026-example}



\appendix
\input{9_Moral_Foundations_Coding_Guide}

\input{Baselines}

\input{10_Baseline_Models_Appendix}

\input{TTC}

\end{document}

%% file: tables/MFRC_f1.tex
\begin{table*}[ht]
\centering
\footnotesize
\setlength{\tabcolsep}{5pt}
\caption{F1 scores (mean ± standard deviation) by moral category on the full MFRC dataset, comparing models under different methods: zero-shot, few-shot prompting, parameter-efficient fine-tuning (PEFT), and full fine-tuning (FFT).}
\label{tab:model_f1}
\begin{tabular}{llccccccc}
\toprule
{Method} & {Model} & {Care} & {Equality} & {Prop} & {Loyalty} & {Authority} & {Purity} & {Thin} \\
\midrule
\multirow{4}{*}{0-shot} 
 & Llama\textsubscript{single} & 0.40\textsubscript{0.004} & 0.29\textsubscript{0.004} & 0.21\textsubscript{0.004} & 0.17\textsubscript{0.004} & 0.20\textsubscript{0.004} & 0.17\textsubscript{0.011} & 0.23\textsubscript{0.006} \\
 & Llama\textsubscript{multi} & 0.28\textsubscript{0.003} & 0.31\textsubscript{0.011} & 0.20\textsubscript{0.017} & 0.14\textsubscript{0.006} & 0.18\textsubscript{0.013} & 0.13\textsubscript{0.017} & 0.01\textsubscript{0.004} \\
 & Ministral\textsubscript{single} & 0.24\textsubscript{0.014} & 0.25\textsubscript{0.010} & 0.15\textsubscript{0.019} & 0.21\textsubscript{0.011} & 0.09\textsubscript{0.008} & 0.10\textsubscript{0.013} & 0.08\textsubscript{0.008} \\
 & Ministral\textsubscript{multi} & 0.41\textsubscript{0.010} & 0.27\textsubscript{0.006} & 0.19\textsubscript{0.005} & 0.17\textsubscript{0.009} & 0.22\textsubscript{0.012} & 0.17\textsubscript{0.009} & 0.20\textsubscript{0.012} \\
\midrule
\multirow{4}{*}{5-shot} 
 & Llama\textsubscript{single} & 0.43\textsubscript{0.005} & 0.21\textsubscript{0.002} & 0.21\textsubscript{0.008} & 0.14\textsubscript{0.002} & 0.20\textsubscript{0.002} & 0.12\textsubscript{0.003} & 0.26\textsubscript{0.005} \\
 & Llama\textsubscript{multi} & 0.33\textsubscript{0.005} & 0.33\textsubscript{0.011} & 0.14\textsubscript{0.022} & 0.18\textsubscript{0.012} & 0.19\textsubscript{0.011} & 0.14\textsubscript{0.020} & 0.01\textsubscript{0.001} \\
 & Ministral\textsubscript{single} & 0.26\textsubscript{0.018} & 0.33\textsubscript{0.009} & 0.13\textsubscript{0.014} & 0.21\textsubscript{0.011} & 0.00\textsubscript{0.002} & 0.21\textsubscript{0.007} & 0.20\textsubscript{0.006} \\
 & Ministral\textsubscript{multi} & 0.44\textsubscript{0.010} & 0.30\textsubscript{0.010} & 0.21\textsubscript{0.013} & 0.18\textsubscript{0.019} & 0.25\textsubscript{0.010} & 0.17\textsubscript{0.017} & 0.15\textsubscript{0.004} \\
\midrule
\multirow{2}{*}{PEFT} 
 & Llama\textsubscript{single} & 0.54\textsubscript{0.012} & 0.51\textsubscript{0.016} & 0.28\textsubscript{0.011} & 0.32\textsubscript{0.007} & 0.34\textsubscript{0.015} & 0.22\textsubscript{0.012} & 0.27\textsubscript{0.024} \\
 & Llama\textsubscript{multi} & 0.53\textsubscript{0.013} & 0.52\textsubscript{0.009} & 0.28\textsubscript{0.024} & 0.40\textsubscript{0.016} & 0.36\textsubscript{0.016} & 0.31\textsubscript{0.037} & 0.25\textsubscript{0.049} \\
\midrule
\multirow{2}{*}{FFT} 
 & BERT\textsubscript{single} & \textbf{0.62\textsubscript{0.020}} & \textbf{0.58\textsubscript{0.030}} & \textbf{0.37\textsubscript{0.040}} & \textbf{0.45\textsubscript{0.040}} & \textbf{0.40\textsubscript{0.050}} & \textbf{0.51\textsubscript{0.070}} & \textbf{0.39\textsubscript{0.020}} \\
 & BERT\textsubscript{multi} & 0.59\textsubscript{0.020} & 0.57\textsubscript{0.030} & 0.31\textsubscript{0.050} & 0.43\textsubscript{0.040} & 0.35\textsubscript{0.040} & 0.48\textsubscript{0.070} & 0.34\textsubscript{0.040} \\
\bottomrule
\end{tabular}
\end{table*}

%% file: tables/New_table.tex
\begin{table*}[!htp]
\centering
\footnotesize
\setlength{\tabcolsep}{5pt}
\caption{Model F1 (mean $\pm$ sd) by Moral Category on the three Subreddit Buckets (French Politics, Everyday, US Politics)}
\label{tab:model_f1_bucket}
\begin{tabular}{llccccccc}
\toprule
{Model} & {Subreddit Bucket} & {Care} & {Equality} & {Prop} & {Loyalty} & {Authority} & {Purity} & {Thin} \\
\midrule
\multirow{3}{*}{BERT\textsubscript{single}} 
  & French Politics & 0.43\textsubscript{0.04}         & 0.59\textsubscript{0.03}           & 0.16\textsubscript{0.05}          & 0.41\textsubscript{0.08}       & 0.20\textsubscript{0.12} & 0.32\textsubscript{0.15} & \textbf{0.41\textsubscript{0.04}} \\
 & Everyday        & \textbf{0.72\textsubscript{0.02}} & \textbf{0.61\textsubscript{0.06}} & 0.37\textsubscript{0.08}            & 0.47\textsubscript{0.08}       & 0.34\textsubscript{0.11} & \textbf{0.51\textsubscript{0.10}} & 0.37\textsubscript{0.05} \\
 & US Politics     & 0.54\textsubscript{0.05}           & 0.58\textsubscript{0.04}          & \textbf{0.42\textsubscript{0.07}} & \textbf{0.51\textsubscript{0.09}}   & \textbf{0.51\textsubscript{0.05}} & 0.37\textsubscript{0.16} & 0.36\textsubscript{0.05} \\
\midrule
\multirow{3}{*}{BERT\textsubscript{multi}}
 & French Politics & 0.33\textsubscript{0.05} & 0.51\textsubscript{0.06} & 0.06\textsubscript{0.07} & 0.33\textsubscript{0.10} & 0.26\textsubscript{0.04} & 0.20\textsubscript{0.16} & 0.37\textsubscript{0.05} \\
 & Everyday & 0.66\textsubscript{0.04} & 0.56\textsubscript{0.07} & 0.34\textsubscript{0.08} & 0.30\textsubscript{0.07} & 0.21\textsubscript{0.10} & 0.45\textsubscript{0.12} & 0.25\textsubscript{0.03} \\
 & US Politics & 0.45\textsubscript{0.07} & 0.55\textsubscript{0.06} & 0.40\textsubscript{0.08} & 0.44\textsubscript{0.11} & 0.43\textsubscript{0.09} & 0.34\textsubscript{0.20} & 0.25\textsubscript{0.07} \\

\bottomrule
\end{tabular}
\end{table*}

%% file: 9_Moral_Foundations_Coding_Guide.tex
\paragraph{Appendix:} Appendix includes the moral foundations coding guide, additional model results, prompts, and Theory Trace Card \citep{karimi2026theory}. For additional material, see the huggingface\footnote{\href{https://huggingface.co/datasets/USC-MOLA-Lab/MFRC}{https://huggingface.co/datasets/USC-MOLA-Lab/MFRC}}.

\section{Moral Foundations Coding Guide} \label{sec:CodingGuide}

Moral expressions in text serve as informationally rich indicators of individuals’ moral values. Whether individuals are signaling their moral beliefs or concerns, framing particular issues or events in moral terms, or expressing a moral emotion, moral expressions are a domain of human language which can inform as to the nature of morality \citep{atari2022}. 
Here, we describe a taxonomy and set of instructions for annotating moral content in natural language, based on Moral Foundations Theory. This taxonomy can be used for the annotation of individual Tweets, Facebook posts, other social media, transcribed speech, and other textual media. 

In this coding guide, we describe the theoretical framework that we rely on to operationalize moral values, Moral Foundations Theory \citep[MFT;][]{haidt2004intuitive,graham2013moral}, describe how moral expressions are annotated, and provide detailed examples and procedures for the process of annotation. \footnote{Please note this guide is based on the the original Moral Foundations Coding Guide which can be found in the Appendix of \cite{hoover2020moral}.}
We follow recent work which expands the original five moral foundations (Care, Fairness, Loyalty, Authority, and Purity) by partitioning Fairness into \enquote{Proportionality} and \enquote{Equality} \citep{atari2022}. 

\subsection{Background: Morality, language analysis, and handling ambiguity}

\subsubsection{Moral Foundations Theory}

Our theoretical framework for annotating morality in language is Moral Foundations Theory  \citep[MFT;][]{haidt2004intuitive,graham2013moral}, a pluralistic, psychological model of moral values.
MFT was developed in order to fill the need of a systematic theory of morality, explaining its evolutionary origins, developmental aspects, and cultural variations. MFT can be viewed as an attempt to specify the psychological mechanisms which allow for intuitive bases of moral judgments as well as moral reasoning. Care, Fairness, Loyalty, Authority, and Purity, according to the original conceptualization of MFT, are five \enquote{foundations} that are conceptualized to have contributed to solving adaptive problems over humans' evolutionary past, and are ubiquitous in current human populations \citep{graham2013moral}.

Each of the five foundations in MFT is conceptualized as having solved different adaptive problems in humans' evolutionary past \citep{haidt2012righteous}. The Care foundation accounts for our nurturing of the young and caring for the infirm. The Fairness foundation accounts for the development of human cooperation, justice, and reciprocity. Loyalty is concerned with coalition-building with ingroup members, Authority is concerned with respecting high-status individuals in social hierarchies, and Purity is about physical cleanliness and spiritual sacredness of objects, humans, and groups. 

One strength of MFT, as formulated by \citeauthor{graham2013moral}, is its openness to new foundations, with the idea that plurality is the most important concept for understanding human morality, and that the specific set of five foundations originally proposed in MFT are just one proposed set of foundations. 
Recently, \citet{atari2022} proposed breaking Fairness into two more narrowly-defined foundations, \enquote{Equality} and \enquote{Proportionality}. Equality describes people's concern with similar outcomes or status (e.g., a violation of Equality is systematic racial inequality, the state of individuals of different races having different access to resources and opportunities). Proportionality describes people's desire for balance between actions and responses. Proportionality concerns are typically centered around the ideas of meritocracy and deservingness. For example, cheaters should be punished, hard workers should be rewarded, and slackers should be excluded relative to the extent of their contribution.

\subsubsection{Moral Foundations in Language}

While explicitly moral language is not common in everyday interactions \citep{atari2022}, moral values, whether implicit or explicit, do play an important role in social functioning \citep{li2021moral}. They influence our judgments and behaviors \citep{ellemers2019psychology,greene2014beyond,haidt2012righteous} and help coordinate complex large-scale cooperation \citep{descioli2013solution,enke2019kinship, dehghani2016purity,purzycki2018cognitive}.

When people express their moral attitudes, emotions, and concerns about people, actions, events, concepts, and ideas, they employ diverse rhetorical strategies \citep{keen2015language}. Often, these strategies rely on words that are explicitly normative, such as \enquote{right,} \enquote{wrong,} \enquote{good,} or \enquote{bad} \citep[which we call ``thin morality''; see][]{atari2022}; however, in many cases, people communicate a moral attitude by communicating the relevance of a moral domain 
(e.g., Care). For example, \enquote{I can’t believe that happened. It’s so harmful!}, \enquote{People should be compassionate,} or \enquote{This decision hurts so many people!} position the discussed entity or topic as either aligned or misaligned with ``good'' morality, by assuming the virtue or desirability of care centered actions, people, or things.

The six moral foundations (i.e., Care, Equality, Proportionality, Loyalty, Authority, and Purity) have a natural mapping to language, which can be instantiated by identifying words which are used by speakers to communicate their attitudes with respect to each moral foundation. In the above examples, speakers' attentiveness to the 
Care foundation is apparent from their usage of the words ``harmful,'' ``compassionate,'' and ``hurt.''
Moral foundations in language were first studied in this way by \citet{graham2009liberals}, which provided the Moral Foundations Dictionary (MFD) \citep[for a detailed discussion of language analysis in moral psychology, see][]{atari2021chapter}. The MFD considers each moral foundation, including the ``vice'' and ``virtue'' poles of each foundation, as a collection of related words, the usage of which indicates a concern with the given foundation. 
In early work, \citet{graham2009liberals} used the MFD to measure differences in moral values sentiment between conservative and liberal sermons. More recently, researchers have shown that moral value annotation based on the MFT taxonomy can fruitfully be applied to a range of applications and domains \citep{dehghani2016purity,sagi2014measuring}. Additionally, extensions and improvements on the initial MFD have been performed by \citet{hopp2021extended} (extended MFD) and \citet{frimer2019moral} (MFD 2.0).  

While the above findings, facilitated by the MFD, 
have shed light on the nature of morality in ``the wild'' by analyzing observational text data, there are reasons to question the use of the dictionary approach for measuring moral phenomena in text. 
There is an implicit assumption that the frequency of certain explicit moral words indicates an underlying concern with the corresponding moral domain. For example, MFD findings articulate implicitly that using moral words implies a higher concern with morality at the individual level. Recently, however, \citet{kennedy2021moral} tested this assumption using responses to the Moral Foundations Questionnaire \citep[MFQ;][]{graham2011mapping} and participants' Facebook status updates. Among other findings, this work established that the existence and size of the relationship between explicitly moral language (i.e., the MFD) and individuals' moral concerns was inconsistent across foundations. Specifically, Care, Fairness, and Purity language had a positive correlation with the corresponding moral domain in language (e.g., high Care concerns implied the usage of Care words), while Authority and Loyalty did not. Moreover, other techniques such as topic modeling, which represent all words (and not just explicitly moral ones) predicted significantly more variance in individual-level moral concerns than did methods based on explicitly moral language.

As far as the present coding guide is concerned, these findings imply that the domain of language which is related to moral concerns is far wider than purely explicit moral language. In terms of the annotation of moral phenomena in text, we will take the view that moral concerns can commonly be communicated without the presence of explicitly moral words, e.g., \enquote{No matter what, it's [my team] forever! 10-0 or 0-10, nothing changes for me} is an expression of loyalty to a sports team without explicitly using words like \enquote{loyalty.}

\subsubsection{Thin Morality}

Not all moral language falls within the scope of the six domains indicated by MFT. In fact, philosophers have denoted two types of morally evaluative language: 

Evaluative terms and concepts are often divided into \enquote{thin} and \enquote{thick}. We don't evaluate actions and persons merely as good or bad, or right or wrong, but also as kind, courageous, tactful, selfish, boorish, and cruel. The latter are examples of thick concepts \dots [which] stand in contrast to those we typically express when we use thin terms such as \textit{right}, \textit{bad}, \textit{permissable}, and \textit{ought} \citep[emphasis in original;][]{vayrynen2016thick}

In our above section on moral concerns in language, we have been discussing thick morality. In our annotation, we will attempt to comprehensively cover all types of moral language by also annotating thin morality.

\subsubsection{Annotator Uncertainty and Authorial Inferences}

There is unavoidable ambiguity that affects text annotation, which is key to understand when conducting annotation-based studies of moral concerns in language. Our approach to ambiguity in this guide is to use background context to inform annotations, but also to report the level of uncertainty for a given annotation. 

The major source of ambiguity is caused by the difficulty of inferring the moral content intended by an author. For example, a social media message might simply state that the author thinks \enquote{Everything that is going on with abortion these days is reprehensible.} In this case, it is clear that this is likely a morally relevant statement, but it is less clear what foundation this statement is relevant to. If we knew that the author was concerned with civil rights, we might assume that the author is concerned about violations of women’s reproductive rights (i.e., an instance of Equality). In contrast, if we knew that the author was a conservative Christian, we might assume that the author was expressing an anti-abortion sentiment, perhaps associated with Purity. 

These ambiguities present considerable challenges for human annotators who must strike an acceptable balance between exploiting often weak signals of moral sentiment while also avoiding unfounded speculation about author's intent. In this guide, we recommend that annotators tend to focus on objective sources of confidence for resolving ambiguities of intent. 
However, since so much in the domain of moral language is not objective (i.e., relying on author assumptions), in this guide we propose the usage of a \enquote{Confidence} label, which is designed to allow annotators to assign a label based on an inference about intent, but to also indicate a lower confidence in the label. %

\subsection{Instructions for Annotators: Annotating moral concerns in language}

Annotating moral concerns in language involves determining whether a given text's author is communicating a moral attitude, emotion, judgment, or moral issues toward particular persons, groups, questions or problems, or event. In this section, we provide instructions for annotators for the identification of moral concerns in text. We emphasize the six domains of moral language based on MFT; detail the \textit{target} and \textit{vice/virtue} components of moral concerns in text; detail instructions for annotators to assign a confidence rating to each annotation; and describe the particular approaches annotators ought to use for different language types (e.g., social media versus transcribed speech).

\subsubsection{Annotation Task}

\begin{figure*}[ht]
    \centering
    \caption{Order of operations for annotating moral values in text: (1) Determine whether one or more moral domains are present (i.e., \textit{thick} morality; (2) Determine whether the text contains \textit{thin} morality; (3) indicate a non-moral text if (1) and (2) categories are not present; and (4) assess confidence in label.}
    \includegraphics[width=0.7\linewidth]{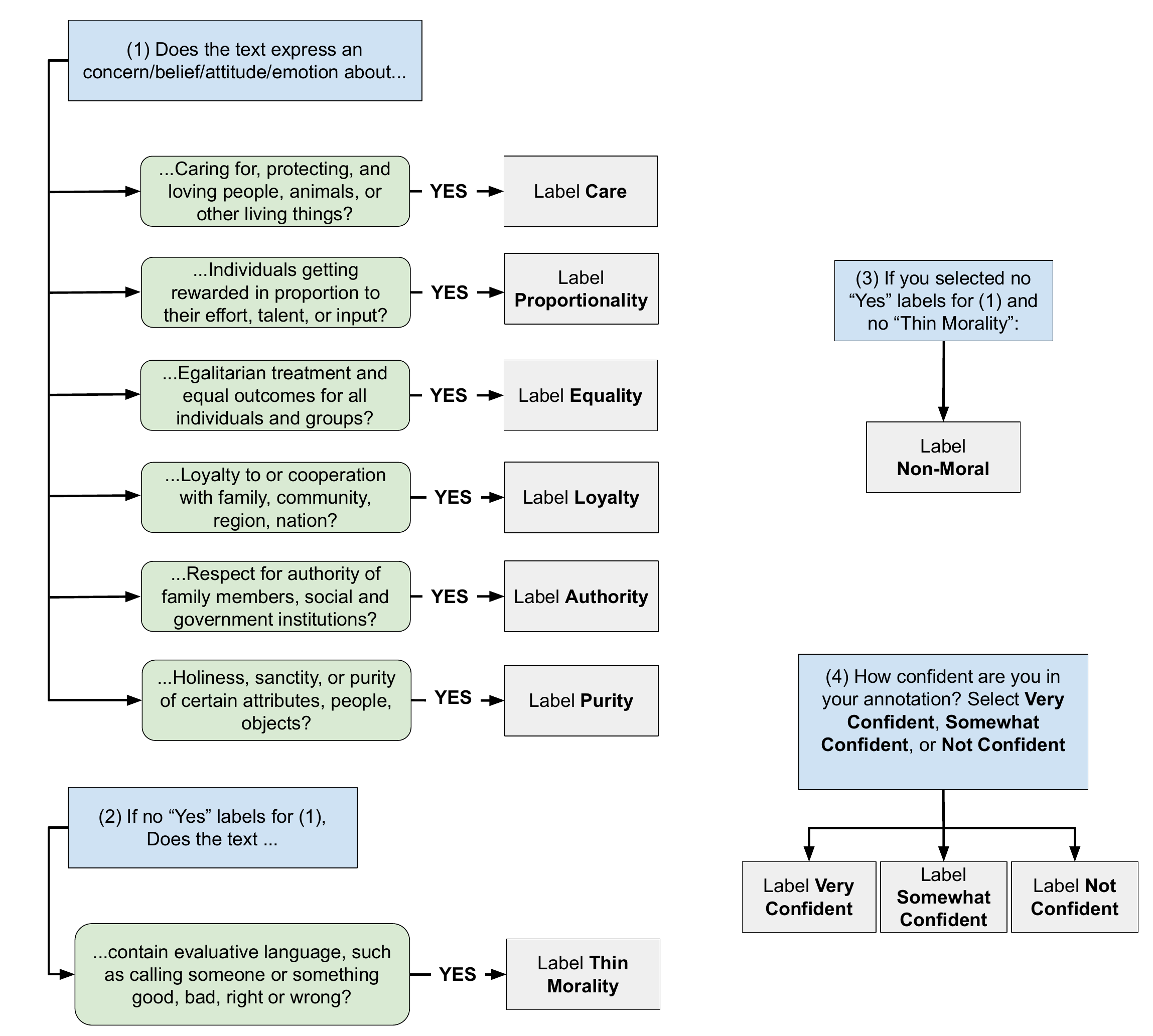}
    \label{fig:flowchart}
\end{figure*}

For moral concern annotation in text, annotators should complete, in order, the four subtasks outlined in Figure~\ref{fig:flowchart}. Below, we will explain each subtask.

\paragraph{Annotating Moral Domains}

We first annotate text by categorizing text into non-mutually exclusive \enquote{domains} of moral concerns, which are the six foundations of MFT.

In Table~\ref{tab:definitions}, we list the six foundations, giving their name, a definition, and an example item from the recently developed MFQ-2 \citep{atari2022}. Items were presented to participants with the prompt, \enquote{Please indicate how well each statement describes you or your opinions.}

\begin{table*}[ht]
\centering
\scriptsize
\caption{Six moral foundations, which map to six domains of moral language}
\label{tab:definitions}
\begin{tabular}{  l  p{7.5cm}  p{5.5cm} }
\toprule
Foundation & Description & Example Item \\\midrule
Care & Intuitions about avoiding emotional and physical damage to another individual. It underlies virtues of kindness, gentleness, and nurturing. &  \textit{I believe that compassion for those who are suffering is one of the most crucial virtues.}  \\\hline
Equality & Intuitions about egalitarian treatment and equal outcome for all individuals and groups. It underlies virtues of social justice and equality. & \textit{The world would be a better place if everyone made the same amount of money}  \\\hline
Proportionality & Intuitions about individuals getting rewarded in proportion to their merit (e.g., effort, talent, or input). It underlies virtues of meritocracy, productiveness, and deservingness. & \textit{The effort a worker puts into a job ought to be reflected in the size of a raise they receive} \\\hline
Loyalty & Intuitions about cooperating with ingroups and competing with outgroups. It underlies virtues of patriotism and self-sacrifice for the group. & \textit{I believe the strength of a sports team comes from the loyalty of its members to each other} \\\hline
Authority & Intuitions about deference toward legitimate authorities and high-status individuals. It underlies virtues of leadership and respect for tradition. & \textit{I think obedience to parents is an important virtue} \\\hline
Purity & Intuitions about avoiding bodily and spiritual contamination and degradation. It underlies virtues of sanctity, nobility, and cleanliness. & \textit{It underlies the widespread idea that the body is a temple that can be desecrated by immoral activities and contaminants (an idea not unique to religious traditions)} 
\\\hline
\end{tabular}
\end{table*}

Even with clarity as to the conceptual domains described by each of the six foundations, it is not straight-forward to map these categories to language. Below, we describe three distinctions, or components, of a voiced moral concern: (1) the explicit/implicit distinction, (2) the concrete versus abstract target distinction, and (3) the vice/virtue (positive/negative) distinction.

\paragraph{Explicit and implicit expressions of moral concerns}

Each of these six concerns can be invoked in language in ways that can be either implicit or explicit. Explicit invocations will use words that map clearly to the domain in question; a sample of such words are given in Table~\ref{tab:mfd_words}.

{\renewcommand{\arraystretch}{1.5}
\begin{table}[ht]
\centering
\scriptsize
\begin{threeparttable}
\setlength{\tabcolsep}{4pt}
\caption{Example words that illustrate each moral language domain including both virtue and vice.}
\label{tab:mfd_words}
\begin{tabularx}{\linewidth}{lXX}
\toprule
Foundation & Virtue & Vice \\
\midrule
Care & compassion\newline kindness & cruel\newline exploit \\
Equality & equal\newline fairly & discriminate\newline injustice \\
Proportionality & proportional\newline deserve & disproportional\newline favoritism \\
Loyalty & collective\newline family & betray\newline disloyal \\
Authority & duty\newline tradition & dissident\newline rioter \\
Purity & sacred\newline chast & sin\newline disgust \\
\bottomrule
\end{tabularx}
\end{threeparttable}
\end{table}
}


The presence of these words in a document/sentence/utterance hints at the presence of a moral expression, but does not necessarily confirm it. An actual invocation of a given moral domain will use these words \textit{in a particular way}. Differentiating what we might call \enquote{moral} uses of these words from \enquote{non-moral} uses is similar to the challenge of \enquote{word sense disambiguation} in Natural Language Processing. Word sense disambiguation acknowledges that words can have multiple uses or \enquote{senses} depending on the context in which they were used. For example, the word \enquote{fairly} can be used in a Care sense, e.g., \enquote{Humanity is we treat every person \textit{fairly}, even when we're threatened,} but can also be used in a non-Care sense, e.g., \enquote{There's \textit{fairly} universal protocol on how to treat anyone who makes dumb decisions}.

Other times, moral concerns can be invoked implicitly (i.e., without using explicitly moral language). This category of moral expression is harder to specify a priori due to its myriad forms; here, annotators should rely on their understanding of the \textit{concepts} underlying each moral domain, as denoted in the flowchart in Figure~\ref{fig:flowchart} and in Table~\ref{tab:definitions}.

For example, it is possible to express a concern about Equality without explicitly naming the Equality domain: \enquote{AT\_USER it's a shame Skin color and beliefs fuel hatred} communicates a concern regarding the fair treatment of people based on skin color or beliefs. Similarly, it is possible to invoke the Loyalty domain without using explicit loyalty words: \enquote{Rep voter [suppression] efforts in Florida a disgrace to Americans my Dad who fought in WWI for freedom \& democracy} describes the virtue of loyalty to American veterans without using the word \enquote{loyalty} (etc.).
The takeaway from these examples is that moral domains can be invoked by understanding the meaning of the text in question, and is not limited to the presence or non-presence of explicitly moral words. 

In summary, there are two types of language which signal the presence of a moral concern:
\begin{enumerate}
    \item Explicitly moral words used in a moral way
    \item Any language used to express the speaker's moral concerns, attitudes, emotions, behavior, or beliefs about some persons, actions, objects, events, ideas, etc.
\end{enumerate}

Next, we detail two sets of distinctions for moral expressions that can help annotators to identify different types of moral language. 

\paragraph{Concrete versus abstract targets of moral expressions}

People often use moral values when they are expressing a judgment about someone or something --- i.e., the \textit{object}. 
The object of a moral judgment can be either \textit{someone} (e.g., a person or a social group) or \textit{something} (e.g., a behavior, an event, an abstract concept, or even a physical object). The object of a moral judgment can be \textit{concrete} (e.g., judging a specific behavior or person) or \textit{abstract} (e.g., judging a general value or opinion). When annotating moral foundations in natural language, we are not, per se, interested in the object of a moral judgments. However, identifying the object of a moral judgment can sometimes help clarify whether the moral judgment in question is related to one of the moral foundations.

\paragraph{Vice and virtue expressions}

The types of moral judgments individuals make about people or things can be either \textit{positive} or \textit{negative}. For example, a person might praise someone for engaging in moral behavior or condemn someone for engaging in immoral behavior. That is, a moral judgment entails a positive or negative evaluation of the object of the moral judgment. 
In broad terms, an expression of virtue communicates that “good should happen” while an expression of vice communicates that \enquote{bad should not happen}—what is \enquote{good} and \enquote{bad} depends, of course, on which moral concern is being evoked.

An evaluation is positive when it calls for moral actions, praises people for moral behavior, or lauds a moral value or opinion. An evaluation is negative when it decries immoral actions, criticizes people for immoral behavior, or condemns an immoral value or opinion. While we are not, per se, interested in distinguishing between positive and negative moral judgements, this distinction can sometimes help clarify what moral foundation is being invoked.

\subsubsection{Annotating Thin Morality}
For our purposes, thin morality is a moral judgment or concern which is voiced without clearly referring to one of the six moral domains. For example, the Tweet \enquote{Why does [he] reply to profane and disrespectful tweets from rude constituents \textbf{he's a good guy}} (emphasis added) makes a statement about the goodness of an individual but does not describe the individual's goodness on account of a particular moral domain. 

We note that Thin morality \textit{is} in fact mutually exclusive with thick morality, and thus if a document is, for example, annotated with Loyalty, it cannot also be labeled as Thin morality. Unlike our definition of thick morality (i.e., the presence of one of the six MFT domains), the presence of thin morality is most often marked by the presence of words (e.g., right, wrong, better, worse, good, bad)

\subsubsection{Annotator Confidence}

After completing annotation of moral domains and Thin morality, and regardless of whether or not Non-Moral was selected, annotators should select one of three confidence labels: Very Confident, Somewhat Confident, and Not Confident. These are fully defined in Table~\ref{tab:confidence_types}.

\begin{table*}[ht]
\centering
\scriptsize
\begin{threeparttable}
\caption{Three possible confidence scores to assign a given annotation, with explanations.}
\begin{tabularx}{\textwidth}{>{\raggedright\arraybackslash}p{1.3cm} X}
\toprule
 & Example Cases \\
\midrule
 & Clearly no moral expression in the text \\
\cmidrule{2-2}
\multirow{3}{=}{Very Confident} 
 & A moral domain is clearly in the text, and it is clear that there are no others \\
\cmidrule{2-2}
 & Multiple domains are clearly in the text, and all are clearly present \\
\midrule
 & No moral expression in the text, but possibility the speaker could be implying a moral concern \\
\cmidrule{2-2}
\multirow{3}{=}{Somewhat Confident}
 & A moral expression in the text, but possibility that the speaker is using sarcasm or similar \\
\cmidrule{2-2}
 & One or more moral domains are clearly present, but at least one is vague or uncertain \\
\midrule
 & No moral label, but with more context it might be possible to establish that the author did intend to communicate a moral concern \\
\cmidrule{2-2}
\multirow{2}{=}{Not Confident}
 & One or more moral labels, but with more context it might be possible to establish that the author did not intend to communicate anything moral \\
\cmidrule{2-2}
 & Two or more moral domains are equally present, but there is no way to resolve either confidently \\
\bottomrule
\end{tabularx}
\label{tab:confidence_types}
\end{threeparttable}
\end{table*}


\subsection{Language Domains and Appropriate Annotation Strategies}

The goal of this coding guide is to be domain-agnostic. That is, rather than a guide for specifically Twitter \citep[e.g.,][]{hoover2020moral} or Facebook data \citep[e.g.,][]{atari2022}, we aim to provide a framework that is flexible enough to guide annotations for any textual domain, including social-media posts, comments on online posts or articles, published text, literary text, historical pieces, and transcribed speech. 

Here, we note particular strategies annotators should take for each language domain. 

\paragraph{Social media (Twitter, Facebook, Instagram, etc.)} 
Social media text is typically short form, containing incomplete sentences, abbreviations, hashtags and `at'-mentions, and hyperlinks. For our annotation, we ask that annotators ignore at-mentions and hyperlinks (including media) and to infer as much as possible from the available context. In some cases, this might include references to current events; for example, the Tweet \enquote{No Social Security number should mean no claim to any benefits or credits. \#takeastand} references the process in the United States whereby a social security number grants access to certain government-provided services. In other cases, abbreviations (e.g., BLM) can be used in ways that add meaningful information. For example, the sentence \enquote{We have endured too much!} might be labeled as Loyalty; however, with the inclusion of the BLM hashtag --- \enquote{We have endured too much! \#BLM}, this might additionally be considered Care or Equality, given that the Black Lives Matter movement is focused on harms and systematic inequalities directed toward Black persons. Annotators are asked to look up unfamiliar abbreviations that occur frequently, or words that seem to have a unique use in a particular online platform or group; however, if the abbreviation is not obvious or frequent, it can be ignored. 
Hashtags, particularly if they are themselves abbreviations, can be used to resolve context. However, they should not be used as the \textit{sole} reason for labeling a document as a given moral domain. For example, the text \enquote{Having some fries with my drink \#Equality} contains a relevant hashtag that does not help to resolve ambiguity, and thus should be ignored. Lastly, hashtags that are used fluidly in a sentence (e.g., \enquote{\#Dreamers play a vital role in our communities}) can be treated as normal words.

In addition to general social media considerations, each platform has specific components that inform annotation strategies. For Twitter, posts are embedded in a networked context with sharing (\enquote{Retweeting}) and conversational components. Currently, we do not support the ability to view conversational context when annotating a Tweet, though this might change in the future. Retweets, marked with \enquote{RT} at the beginning of the Tweet, and Quote Tweets, marked by a preceding remark followed by the main Tweet in quotes, should be viewed as endorsing the message contained in the original tweet, and annotated accordingly. Lastly, most of the considerations that apply to Twitter apply to other short-form social media, such as Instagram.

For Facebook, posts can be longer (i.e., multi-sentence or multi-paragraph), requiring more time than Tweets and other short form text. However, Facebook posts are typically less ambiguous than Tweets, as they contain more context. The moral concerns voiced in a Facebook posts will likely be contained in one or two select sentences. Additionally, it is more likely for a Facebook post to contain multiple moral concerns than it shorter media like Twitter.

    
\paragraph{Online comments (Reddit)}
The distinguishing characteristic of online comment language is its referencing of original posts. For example, a comment on a Reddit post in the forum \enquote{Am I the Asshole} will be making judgments or comments about a post in which the original author explained a personal story, asking anonymous ethical judges to pronounce judgment on the individuals in the story. For our purposes, we will not have annotators read original posts in large part due to the length of original posts. Instead of relying on this context, we will ask annotators to label comments using only the language contained in the given comment. This has limitations with regard to resolving ambiguities, and thus annotators should take care to report annotator confidence when additional context would be needed to label an ambiguous comment.

\paragraph{Transcribed speech}

Spoken language is altogether different from written language, due to the difference between spontaneous conversation and the premeditated nature of written text. Transcriptions of speech capture all the artifacts of speech, including \enquote{er} and \enquote{ah} sounds, short sentences, and incomplete sentences. Also, the types of moral concern voiced in spoken language tends to be more concrete than abstract. The following examples illustrate the types of language in (transcribed) spoken text: \enquote{That's good. I'm happy you're taking care of you [\textit{sic}] mom} (Care); \enquote{I will never leave your side} (Loyalty); \enquote{I'm your dad! You need to respect me!} (Authority); \enquote{John is a good man} (Thin Morality). 

\paragraph{Published text/articles}
Lastly, published text or articles, whether sampled at the sentence, paragraph, or document level, require annotators to read carefully and to consider as much external context as possible. Similarly to Facebook posts, annotators should attempt to identify sentences or sequences of sentences that contain a give moral concern. Additionally, given the nature of published text and articles, more is known about the speaker, or at least the speaker's objectives, which might be to persuade readers about a certain point or to describe a story or event. This context can be used to resolve ambiguities in text.




\subsection{Examples}

\subsubsection{\textit{\textbf{Examples of each Label from the MFTC}}}

See Table \ref{tab:examples_general}.

\begin{table*}[htp]
    \centering
    \scriptsize
    \begin{threeparttable}
    \caption{Examples from MFTC of each label in our taxonomy}
    \begin{tabular}{p{8.0cm}   p{1.5cm} p{6.5cm}}
    \hline
    Example & Foundation & Explanation \\
    \hline
       
       please remember to watch for frightened lost injured pets & Care & Asks others to care for (virtue) injured pets (object) \\
       If hurricane Sandy hurts anyone I love She will be cunt punted & Care & Threatens violence (vice) if loved ones are hurt (object/event) \\
       I'm rooting for equality \#iamAME \#BlackLivesMatter \#AllLivesMatter & Equality & Positive support (virtue) for equality (object) \\
      Why is no one worried about disenfranchisement caused by lack of electricity From a NJ voter election voting rights no electric Sandy & Equality & Expressing about worry about inequality (vice) due to lack of electricity (object) \\
      Winning fair share of Wealth \& Power will be key to any lasting change & Proportionality & Calls for fair share (virtue) of wealth and power (object) \\
      USER should have asked if Springsteen was going to stop taking advantage of farmer loopholes and pay his fair share of taxes & Proportionality & Implies negative attitude (vice) toward someone not paying their fair share (object) \\
      Solidarity Sunday. \#blacklivesmatter. \#icantbreathe & Loyalty & Expresses solidarity (virtue) with the Black Lives Matter movement \\
      @LindseyGrahamSC Be a true patriot \& speak up & Loyalty & Telling someone (object) to be a true patriot (virtue) \\
      No to illegal immigrants-they need to follow the process, obey the law @realDonaldTrump \#EndDACA & Authority & Expresses that people should obey (virtue) the US president (object) \\
      @realDonaldTrump ``If you love me, obey my commandments.'' -Jesus John 14:15 & Authority & Telling some (virtue) to obey God's commandments (object) \\
      Glad to see a reformation going on to restore sanctity of free speech & Purity & Praising (virtue) a restoration of the sanctity of free speech (object) \\
      It's absolutely disgusting how every retailer exploits a serious storm situation by peddling their crap & Purity & Expression of disgust (vice) regarding exploiting a storm situation to sell wares (object) \\
\bottomrule
\end{tabular}
\label{tab:examples_general}
\end{threeparttable}
\end{table*}

\subsubsection{Examples of Thin Morality}

See Table \ref{tab:examples_thin}.

{\renewcommand{\arraystretch}{1.2}
\begin{table*}[ht]
\scriptsize
\centering
\begin{threeparttable}
\caption{Examples of Thin Morality as well as Non-Moral examples.}
\begin{tabular}{p{7.0cm}   p{1.5cm} p{7cm}}
\toprule
     Example & Label & Notes \\\midrule
     John is a good man. & Thin Morality & -\\
     Yes I think that's correct. & Non-Moral & Agreeing with someone, not expressing some moral evaluation. \\
     What he did was absolutely wrong. Unacceptable! & Thin Morality & -\\
     Mother Theresa's goodness won her a Nobel Prize. & Thin Morality & Praising on account of goodness. \\
     I have no idea what to say. Hmmmm... & Non-Moral & -\\
     \bottomrule
\end{tabular}
\label{tab:examples_thin}
\end{threeparttable}
\end{table*}
}

\subsubsection{Examples of Equality and Proportionality}

Special attention is given in this coding guide to the annotation of Equality and Proportionality. Table~\ref{tab:updated_labels_examples}, we give illustrative examples of Tweets from the MFTC which were previously annotated as Fairness(Vice/Virtue), but in the present coding guide are either Equality or Proportionality.

\begin{table*}[!ht]
\centering
\scriptsize
\begin{threeparttable}
\caption{Examples of changes in labeling for the new, six-foundation taxonomy}
\label{tab:updated_labels_examples}
\begin{tabular}{p{10.0cm}   p{1.8cm} p{1.8cm}}
\toprule
    Text & Original Labels & New Labels \\\hline
    \#AllLivesMatter is a cop out. A convenient way to dismiss oppression and inequality in this country. & Harm, Cheating & Inequality \\
    If you choose to be a police officer, you have a responsibility to uphold justice and treat everyone equal. \#AllLivesMatter & Fairness & Equality \\
    RT @CNN: "No justice, no peace." Crowds protest the death of \#FreddieGray in \#Baltimore & Non-Moral & Proportionality \\
    To hear the police union president say the officers did nothing wrong breaks my heart. I can't even use the angry emotion. & Cheating & Proportionality\\
    RT @chescaleigh: talking about injustice shouldn’t upset you. the injustice should. \#BlackLivesMatter & Cheating & Propotionality \\
    I cancelled my direct debit and I'm going to refuse to pay! This is fraud by O2 & Cheating & Disproportionality \\
\bottomrule
\end{tabular}
\end{threeparttable}
\end{table*}



%% file: Baselines.tex
\section{Personalization} \label{sec:personalization}

In this section, we focus on disaggregated performance and present results on each annotator, or the personalization capabilities that the dataset affords. Results are aggregated across labels, and therefore multi-label metrics are presented: Sample Semantic F1 (\citealp{chochlakis2025semantic}; the same similarities as in the original paper are used for evaluation), Jaccard score, Macro F1 and Micro F1~\cite{fujino2008multi, loza2023tree}. In this section, we use Llama3-8B and Llama3-70B~\cite{dubey2024llama}, Qwen3-30B~\cite{yang2025qwen3}, and GPT-OSS-20B~\cite{agarwal2025gpt}.

We present results in the few-shot setting (10, 20, and 30 shots), as well as more explicit prompting that incorporates the annotation manual to precisely define terms and derivation process. The prompts are shown in Table~\ref{tab:prompt-example}. Few-shot learning attempts to personalize the predictions to the annotator's choices, whereas definitions (which are 0-shot for GPT-OSS and 1-shot otherwise) leverage the general semantics to ground the predictions to the manual.

Thorough results, including standard deviations, are presented in Tables~\ref{tab:mfrc-semantic-annotators}, \ref{tab:mfrc-js-annotators}, \ref{tab:mfrc-macro-annotators}, \ref{tab:mfrc-micro-annotators} respectively, both for the aggregate and for each annotator.
Overall  we see that reasoning models perform best with the definitions, while the rest achieve better performance with few-shot learning instead. We see that reasoning models outperform the rest (irrespective of setting) in micro and macro F1, which are more sensitive to thin rationality, and instruction-tuned models better capture the morality of comments with respect to Semantic F1 and Jaccard Score. Depending on the metric, therefore, personalization models perform better or worse than models guessing predictions based on the manual, proving the personalization study inconclusive. We note that sample-based evaluations, like Sample Semantic F1 and Jaccard Score, might overestimate the performance of models given the sparsity of the labels.

Similar to previous evaluations~\cite{chochlakis2025aggregation}, we also find that the aggregate tends to have lower performance compared to annotators when measuring performance with micro and macro F1.

\begin{table*}[!ht]
  \centering
  \caption{Sample Semantic F1 comparison across annotators.}
  \label{tab:mfrc-semantic-annotators}
  \begin{adjustbox}{width=0.7\textwidth}
  \begin{tabular}{llccccccc}
    \toprule
    Method & Model & Aggregate & 0 & 1 & 2 & 3 & 4 & 5 \\
    \midrule

    \multirow{3}{*}{10-shot} &
    Qwen3 30B Instruct & 0.72\textsubscript{0.015} & 0.70\textsubscript{0.010} & 0.64\textsubscript{0.011} & 0.69\textsubscript{0.007} & 0.73\textsubscript{0.011} & 0.61\textsubscript{0.007} & 0.65\textsubscript{0.019} \\

    &Llama3 8B Instruct & 0.67\textsubscript{0.009} & 0.65\textsubscript{0.009} & 0.58\textsubscript{0.012} & 0.67\textsubscript{0.026} & 0.72\textsubscript{0.009} & 0.61\textsubscript{0.016} & 0.68\textsubscript{0.032} \\

    &GPT-OSS 20B & 0.62\textsubscript{0.002} & 0.57\textsubscript{0.017} & 0.61\textsubscript{0.008} & 0.60\textsubscript{0.007} & 0.63\textsubscript{0.004} & 0.59\textsubscript{0.003} & 0.72\textsubscript{0.035} \\

    \midrule
    
    \multirow{3}{*}{20-shot} &
    Qwen3 30B Instruct & 0.75\textsubscript{0.014} & 0.72\textsubscript{0.029} & 0.67\textsubscript{0.001} & 0.69\textsubscript{0.019} & 0.77\textsubscript{0.013} & 0.63\textsubscript{0.015} & 0.66\textsubscript{0.049} \\

    &Llama3 8B Instruct & 0.72\textsubscript{0.006} & 0.69\textsubscript{0.017} & 0.62\textsubscript{0.006} & 0.70\textsubscript{0.015} & 0.75\textsubscript{0.009} & 0.63\textsubscript{0.007} & 0.70\textsubscript{0.020} \\

    &GPT-OSS 20B & 0.64\textsubscript{0.012} & 0.61\textsubscript{0.018} & 0.62\textsubscript{0.002} & 0.66\textsubscript{0.002} & 0.66\textsubscript{0.013} & 0.60\textsubscript{0.009} & 0.69\textsubscript{0.006} \\

    \midrule
    \multirow{2}{*}{30-shot} &
    Qwen3 30B Instruct & 0.76\textsubscript{0.004} & 0.75\textsubscript{0.009} & 0.68\textsubscript{0.016} & 0.73\textsubscript{0.010} & 0.78\textsubscript{0.006} & 0.62\textsubscript{0.010} & 0.71\textsubscript{0.030} \\

    &Llama3 8B Instruct & 0.74\textsubscript{0.010} & 0.72\textsubscript{0.009} & 0.66\textsubscript{0.007} & 0.72\textsubscript{0.008} & 0.77\textsubscript{0.008} & 0.64\textsubscript{0.003} & 0.70\textsubscript{0.026} \\

    \midrule
    \multirow{4}{*}{w/ defs} & 
    Qwen3 30B Instruct & 0.54\textsubscript{0.010} & 0.52\textsubscript{0.011} & 0.61\textsubscript{0.009} & 0.54\textsubscript{0.008} & 0.56\textsubscript{0.010} & 0.56\textsubscript{0.013} & 0.63\textsubscript{0.033} \\
    
    &Llama3 8B Instruct & 0.62\textsubscript{0.009} & 0.58\textsubscript{0.001} & 0.61\textsubscript{0.009} & 0.59\textsubscript{0.008} & 0.63\textsubscript{0.011} & 0.59\textsubscript{0.010} & 0.62\textsubscript{0.048} \\
    
    &Llama3 70B Instruct & 0.47\textsubscript{0.011} & 0.45\textsubscript{0.013} & 0.55\textsubscript{0.014} & 0.46\textsubscript{0.006} & 0.47\textsubscript{0.011} & 0.52\textsubscript{0.006} & 0.67\textsubscript{0.013} \\
    
    &GPT-OSS 20B & 0.68\textsubscript{0.011} & 0.66\textsubscript{0.005} & 0.66\textsubscript{0.005} & 0.68\textsubscript{0.008} & 0.70\textsubscript{0.008} & 0.62\textsubscript{0.013} & 0.66\textsubscript{0.023} \\
    
    \bottomrule
  \end{tabular}
  \end{adjustbox}
\end{table*}

\begin{table*}[!ht]
  \centering
  \caption{Jaccard score comparison across annotators.}
  \label{tab:mfrc-js-annotators}
  \begin{adjustbox}{width=0.7\textwidth}
  \begin{tabular}{llccccccc}
    \toprule
    Method & Model & Aggregate & 0 & 1 & 2 & 3 & 4 & 5 \\
    \midrule
    \multirow{3}{*}{10-shot} & Qwen3 30B Instruct & 0.67\textsubscript{0.013} & 0.64\textsubscript{0.006} & 0.50\textsubscript{0.011} & 0.63\textsubscript{0.002} & 0.68\textsubscript{0.010} & 0.48\textsubscript{0.014} & 0.42\textsubscript{0.024} \\
     & Llama3 8B Instruct & 0.64\textsubscript{0.009} & 0.59\textsubscript{0.012} & 0.44\textsubscript{0.012} & 0.60\textsubscript{0.023} & 0.66\textsubscript{0.004} & 0.47\textsubscript{0.012} & 0.48\textsubscript{0.026} \\
     & GPT-OSS 20B & 0.54\textsubscript{0.001} & 0.49\textsubscript{0.026} & 0.47\textsubscript{0.006} & 0.51\textsubscript{0.001} & 0.55\textsubscript{0.008} & 0.44\textsubscript{0.001} & 0.49\textsubscript{0.015} \\

    \midrule
    \multirow{3}{*}{20-shot} & Qwen3 30B Instruct & 0.69\textsubscript{0.015} & 0.66\textsubscript{0.026} & 0.55\textsubscript{0.006} & 0.64\textsubscript{0.021} & 0.72\textsubscript{0.012} & 0.50\textsubscript{0.008} & 0.47\textsubscript{0.038} \\
     & Llama3 8B Instruct & 0.67\textsubscript{0.006} & 0.64\textsubscript{0.017} & 0.50\textsubscript{0.009} & 0.63\textsubscript{0.011} & 0.70\textsubscript{0.006} & 0.51\textsubscript{0.005} & 0.52\textsubscript{0.007} \\
     & GPT-OSS 20B & 0.58\textsubscript{0.012} & 0.54\textsubscript{0.019} & 0.50\textsubscript{0.009} & 0.57\textsubscript{0.001} & 0.59\textsubscript{0.018} & 0.45\textsubscript{0.008} & 0.46\textsubscript{0.005} \\

    \midrule
    \multirow{2}{*}{30-shot} & Qwen3 30B Instruct & 0.72\textsubscript{0.001} & 0.70\textsubscript{0.006} & 0.57\textsubscript{0.015} & 0.68\textsubscript{0.010} & 0.74\textsubscript{0.009} & 0.51\textsubscript{0.004} & 0.51\textsubscript{0.037} \\
     & Llama3 8B Instruct & 0.69\textsubscript{0.007} & 0.67\textsubscript{0.009} & 0.55\textsubscript{0.009} & 0.66\textsubscript{0.011} & 0.72\textsubscript{0.007} & 0.53\textsubscript{0.009} & 0.52\textsubscript{0.037} \\

    \midrule
    \multirow{4}{*}{w/ defs} & Qwen3 30B Instruct & 0.46\textsubscript{0.010} & 0.44\textsubscript{0.011} & 0.44\textsubscript{0.010} & 0.44\textsubscript{0.006} & 0.48\textsubscript{0.011} & 0.39\textsubscript{0.011} & 0.42\textsubscript{0.028} \\
     & Llama3 8B Instruct & 0.54\textsubscript{0.007} & 0.50\textsubscript{0.002} & 0.45\textsubscript{0.008} & 0.50\textsubscript{0.005} & 0.55\textsubscript{0.011} & 0.45\textsubscript{0.010} & 0.41\textsubscript{0.050} \\
     & Llama3 70B Instruct & 0.40\textsubscript{0.012} & 0.37\textsubscript{0.012} & 0.39\textsubscript{0.009} & 0.36\textsubscript{0.006} & 0.40\textsubscript{0.010} & 0.34\textsubscript{0.008} & 0.40\textsubscript{0.005} \\
     & GPT-OSS 20B & 0.63\textsubscript{0.012} & 0.60\textsubscript{0.008} & 0.57\textsubscript{0.006} & 0.61\textsubscript{0.010} & 0.65\textsubscript{0.005} & 0.49\textsubscript{0.009} & 0.50\textsubscript{0.025} \\

    \bottomrule
  \end{tabular}
  \end{adjustbox}
\end{table*}

\begin{table*}[!ht]
  \centering
  \caption{Macro F1 comparison across annotators.}
  \label{tab:mfrc-macro-annotators}
  \begin{adjustbox}{width=0.7\textwidth}
  \begin{tabular}{llccccccc}
    \toprule
    Method & Model & Aggregate & 0 & 1 & 2 & 3 & 4 & 5 \\
    \midrule
    \multirow{3}{*}{10-shot} & Qwen3 30B Instruct & 0.27\textsubscript{0.019} & 0.21\textsubscript{0.013} & 0.31\textsubscript{0.004} & 0.27\textsubscript{0.027} & 0.26\textsubscript{0.017} & 0.31\textsubscript{0.028} & 0.26\textsubscript{0.024} \\
     & Llama3 8B Instruct & 0.25\textsubscript{0.016} & 0.19\textsubscript{0.019} & 0.25\textsubscript{0.008} & 0.21\textsubscript{0.027} & 0.21\textsubscript{0.014} & 0.32\textsubscript{0.018} & 0.32\textsubscript{0.038} \\
     & GPT-OSS 20B & 0.22\textsubscript{0.002} & 0.20\textsubscript{0.019} & 0.31\textsubscript{0.002} & 0.21\textsubscript{0.024} & 0.20\textsubscript{0.016} & 0.32\textsubscript{0.004} & 0.29\textsubscript{0.039} \\

    \midrule
    \multirow{3}{*}{20-shot} & Qwen3 30B Instruct & 0.28\textsubscript{0.015} & 0.21\textsubscript{0.018} & 0.34\textsubscript{0.005} & 0.26\textsubscript{0.031} & 0.29\textsubscript{0.008} & 0.33\textsubscript{0.008} & 0.27\textsubscript{0.045} \\
     & Llama3 8B Instruct & 0.25\textsubscript{0.022} & 0.20\textsubscript{0.014} & 0.26\textsubscript{0.017} & 0.17\textsubscript{0.012} & 0.26\textsubscript{0.012} & 0.33\textsubscript{0.007} & 0.36\textsubscript{0.027} \\
     & GPT-OSS 20B & 0.25\textsubscript{0.017} & 0.23\textsubscript{0.015} & 0.32\textsubscript{0.029} & 0.25\textsubscript{0.005} & 0.24\textsubscript{0.016} & 0.32\textsubscript{0.015} & 0.31\textsubscript{0.007} \\

    \midrule
    \multirow{2}{*}{30-shot} & Qwen3 30B Instruct & 0.30\textsubscript{0.023} & 0.25\textsubscript{0.007} & 0.33\textsubscript{0.014} & 0.29\textsubscript{0.002} & 0.27\textsubscript{0.026} & 0.32\textsubscript{0.008} & 0.26\textsubscript{0.014} \\
     & Llama3 8B Instruct & 0.25\textsubscript{0.012} & 0.22\textsubscript{0.013} & 0.29\textsubscript{0.012} & 0.22\textsubscript{0.007} & 0.23\textsubscript{0.033} & 0.35\textsubscript{0.019} & 0.35\textsubscript{0.061} \\

    \midrule
    \multirow{4}{*}{w/ defs} & Qwen3 30B Instruct & 0.25\textsubscript{0.006} & 0.23\textsubscript{0.009} & 0.31\textsubscript{0.010} & 0.25\textsubscript{0.006} & 0.25\textsubscript{0.005} & 0.32\textsubscript{0.009} & 0.30\textsubscript{0.018} \\
     & Llama3 8B Instruct & 0.23\textsubscript{0.005} & 0.20\textsubscript{0.005} & 0.29\textsubscript{0.009} & 0.22\textsubscript{0.005} & 0.20\textsubscript{0.012} & 0.30\textsubscript{0.006} & 0.27\textsubscript{0.034} \\
     & Llama3 70B Instruct & 0.26\textsubscript{0.011} & 0.20\textsubscript{0.010} & 0.31\textsubscript{0.010} & 0.25\textsubscript{0.013} & 0.23\textsubscript{0.011} & 0.32\textsubscript{0.006} & 0.36\textsubscript{0.012} \\
     & GPT-OSS 20B & 0.35\textsubscript{0.027} & 0.32\textsubscript{0.018} & 0.39\textsubscript{0.021} & 0.31\textsubscript{0.019} & 0.30\textsubscript{0.011} & 0.28\textsubscript{0.010} & 0.35\textsubscript{0.053} \\

    \bottomrule
  \end{tabular}
  \end{adjustbox}
\end{table*}

\begin{table*}[!ht]
  \centering
  \caption{Micro F1 comparison across annotators.}
  \label{tab:mfrc-micro-annotators}
  \begin{adjustbox}{width=0.7\textwidth}
  \begin{tabular}{llccccccc}
    \toprule
    Method & Model & Aggregate & 0 & 1 & 2 & 3 & 4 & 5 \\
    \midrule
    \multirow{3}{*}{10-shot} & Qwen3 30B Instruct & 0.32\textsubscript{0.019} & 0.26\textsubscript{0.013} & 0.35\textsubscript{0.011} & 0.32\textsubscript{0.021} & 0.30\textsubscript{0.015} & 0.36\textsubscript{0.028} & 0.31\textsubscript{0.024} \\
     & Llama3 8B Instruct & 0.29\textsubscript{0.012} & 0.24\textsubscript{0.011} & 0.29\textsubscript{0.009} & 0.24\textsubscript{0.016} & 0.25\textsubscript{0.008} & 0.35\textsubscript{0.018} & 0.39\textsubscript{0.048} \\
     & GPT-OSS 20B & 0.24\textsubscript{0.004} & 0.24\textsubscript{0.020} & 0.35\textsubscript{0.001} & 0.22\textsubscript{0.021} & 0.22\textsubscript{0.018} & 0.34\textsubscript{0.002} & 0.31\textsubscript{0.026} \\

    \midrule
    \multirow{3}{*}{20-shot} & Qwen3 30B Instruct & 0.34\textsubscript{0.020} & 0.28\textsubscript{0.027} & 0.37\textsubscript{0.002} & 0.32\textsubscript{0.029} & 0.35\textsubscript{0.013} & 0.38\textsubscript{0.004} & 0.33\textsubscript{0.040} \\
     & Llama3 8B Instruct & 0.31\textsubscript{0.014} & 0.26\textsubscript{0.013} & 0.31\textsubscript{0.013} & 0.24\textsubscript{0.015} & 0.29\textsubscript{0.012} & 0.38\textsubscript{0.010} & 0.41\textsubscript{0.020} \\
     & GPT-OSS 20B & 0.29\textsubscript{0.018} & 0.27\textsubscript{0.017} & 0.38\textsubscript{0.013} & 0.28\textsubscript{0.003} & 0.27\textsubscript{0.013} & 0.34\textsubscript{0.012} & 0.31\textsubscript{0.009} \\

    \midrule
    \multirow{2}{*}{30-shot} & Qwen3 30B Instruct & 0.36\textsubscript{0.009} & 0.31\textsubscript{0.016} & 0.37\textsubscript{0.009} & 0.36\textsubscript{0.016} & 0.34\textsubscript{0.013} & 0.37\textsubscript{0.006} & 0.34\textsubscript{0.017} \\
     & Llama3 8B Instruct & 0.31\textsubscript{0.009} & 0.28\textsubscript{0.015} & 0.36\textsubscript{0.016} & 0.28\textsubscript{0.015} & 0.29\textsubscript{0.019} & 0.39\textsubscript{0.015} & 0.42\textsubscript{0.047} \\

    \midrule
    \multirow{4}{*}{w/ defs} & Qwen3 30B Instruct & 0.25\textsubscript{0.006} & 0.24\textsubscript{0.009} & 0.32\textsubscript{0.009} & 0.25\textsubscript{0.007} & 0.23\textsubscript{0.005} & 0.33\textsubscript{0.010} & 0.35\textsubscript{0.021} \\
     & Llama3 8B Instruct & 0.23\textsubscript{0.003} & 0.20\textsubscript{0.005} & 0.30\textsubscript{0.008} & 0.22\textsubscript{0.004} & 0.21\textsubscript{0.011} & 0.32\textsubscript{0.004} & 0.28\textsubscript{0.036} \\
     & Llama3 70B Instruct & 0.27\textsubscript{0.009} & 0.24\textsubscript{0.013} & 0.37\textsubscript{0.008} & 0.24\textsubscript{0.007} & 0.25\textsubscript{0.005} & 0.34\textsubscript{0.007} & 0.33\textsubscript{0.011} \\
     & GPT-OSS 20B & 0.40\textsubscript{0.021} & 0.37\textsubscript{0.015} & 0.44\textsubscript{0.018} & 0.34\textsubscript{0.018} & 0.37\textsubscript{0.011} & 0.32\textsubscript{0.008} & 0.34\textsubscript{0.043} \\

    \bottomrule
  \end{tabular}
  \end{adjustbox}
\end{table*}

\begin{table*}[!ht]
    \centering
    \scriptsize
    \renewcommand{\arraystretch}{1}
    \begin{tabular}[t]{@{}p{\textwidth}@{}}

    \toprule

    \textbf{Few-shot prompt} \\
    \narrowbotc{Classify the following inputs into none, one, or multiple the following moral foundations per input: authority, care, equality, loyalty, proportionality and purity.\\\\Input: `Or maybe her 'picker' is broken. This guy seems really desperate to justify things that don't need justification, and by doing so sets my spidey sense a'tingling.`\\\{"label": ["none"]\}} \\\\

    \textbf{Definition prompt} \\
    \narrowbotc{Assign ZERO OR MORE labels from this set (multi-label allowed): authority, care, equality, loyalty, proportionality and purity\\\\KEY IDEA: WHAT “THICK MORALITY” MEANS HERE\\“Thick morality” means a moral judgment that is tied to a specific kind of moral reason, i.e., it clearly maps onto at least one of the six foundation domains below.\\So: thick morality = (moral evaluation) + (domain-specific basis).\\If the text is moral but the basis is not identifiable (just generic good/bad/right/wrong), that is NOT thick morality for this task.\\\\CORE MEANINGS (THE DOMAIN-SPECIFIC BASES)\\- Authority: respect for legitimate roles, rules, and tradition; duty, obedience, order, insubordination, disrespect.\\- Care: preventing or responding to suffering; compassion, cruelty, harm, protection, neglect.\\- Equality: equal treatment and equal standing; discrimination, oppression, civil rights, unfair exclusion.\\- Loyalty: commitment to an in-group; solidarity, patriotism, betrayal, disloyalty, “us vs them” obligations.\\- Proportionality: merit-based fairness; deservingness, earned reward, freeloading, cheating, corruption, unfair advantage.\\- Purity: sanctity and contamination concerns; cleanliness, chastity, disgust, “degrading” acts, sacred/profane framing.\\\\DECISION PROCEDURE (DO THIS IN ORDER)\\Step 1: Check for thick morality.\\Does the text express a moral concern/judgment AND is the moral basis clearly one or more of the six domains?\\- If yes: output all applicable domain labels and STOP.\\- If no: continue.\\\\Step 2: Otherwise output no-label.\\This includes:\\- generic moral language with no identifiable domain (e.g., “that’s wrong”, “immoral”, “people should be better” with no clear reason)\\- non-moral content (facts, preferences, jokes, logistics, neutral descriptions)\\\\WHAT COUNTS AS “EXPRESSES A DOMAIN”\\A domain is present if the text contains a moral evaluation, emotion, norm, obligation, blame/praise, or call-to-action grounded in that domain.\\- Virtue or vice both count (praise or condemnation).\\- Multiple domains can co-occur; label all that clearly apply.\\\\AVOID THESE FAILURE MODES\\- Mind-reading: Don’t invent intent or moral meaning that isn’t supported by the text. If it’s underdetermined, don’t guess.\\- Keyword anchoring: Moral concerns can be implicit. Moral-sounding words can be non-moral. Label from meaning, not wordlists.\\- Thin vs domain mix-up: If a foundation domain fits, use it. Generic “right/wrong” with no clear domain is different and only applies when no domain fits.\\- Skipping the procedure: Always check domains → generic morality → non-moral. Don’t jump straight to non-moral.\\- Uncertainty blind spots: If missing context or sarcasm could flip the stance or the domain, don’t overcommit.\\- Multi-domain handling: Domains can co-occur. Label all that clearly apply. If multiple domains are equally plausible and you can’t resolve, don’t force it.\\\\Input: `Make no mistake, Le Pen was a friend to jews and the gay community.\\Her loss will lead to an increase in murders and hate-crime against those communities, inevitably. `\\\{"label": ["care", "equality"]\}}
    
    \end{tabular}
    \caption{Example prompts}
    \label{tab:prompt-example}
\end{table*}

%% file: 10_Baseline_Models_Appendix.tex
\section{Additional Baseline Model Results}\label{sec:BaselineResults}
See Tables \ref{tab:model_precision} - \ref{tab:model_recall}.
\begin{table*}[h!]
\centering
\footnotesize
\caption{Model Precision (mean $\pm$ sd) by Moral Category on the full MFRC dataset.}
\label{tab:model_precision}
\resizebox{.7\textwidth}{!}{%
\begin{tabular}{lccccccc}
\toprule
{Model} & {Care} & {Equality} & {Prop} & {Loyalty} & {Authority} & {Purity} & {Thin} \\
\midrule
BERT & 0.58 (.05) & 0.60 (.05) & 0.44 (.07) & 0.48 (.05) & 0.39 (.04) & \textbf{0.69 (.12)} & 0.36 (.03) \\
ML-BERT & 0.61 (.02) & 0.61 (.04) & 0.40 (.06) & 0.54 (.08) & 0.42 (.04) & 0.65 (.10) & 0.41 (.06) \\
Llama-0-shot-single & 0.41 (.01) & 0.29 (.01) & 0.23 (.01) & 0.18 (.01) & 0.20 (.01) & 0.18 (.01) & 0.24 (.01) \\
Llama-0-shot-multi & 0.29 (.01) & 0.32 (.01) & 0.21 (.02) & 0.15 (.01) & 0.19 (.01) & 0.14 (.02) & 0.02 (.01) \\
Ministral-0-shot-single & 0.25 (.01) & 0.26 (.01) & 0.16 (.02) & 0.22 (.01) & 0.10 (.01) & 0.11 (.01) & 0.09 (.01) \\
Ministral-0-shot-multi & 0.42 (.01) & 0.28 (.01) & 0.20 (.01) & 0.18 (.01) & 0.23 (.01) & 0.18 (.01) & 0.21 (.01) \\
Llama-5-shot-single & 0.44 (.01) & 0.22 (.01) & 0.22 (.01) & 0.15 (.01) & 0.21 (.01) & 0.13 (.01) & 0.27 (.01) \\
Llama-5-shot-multi & 0.34 (.01) & 0.34 (.01) & 0.15 (.02) & 0.19 (.01) & 0.20 (.01) & 0.15 (.02) & 0.02 (.01) \\
Ministral-5-shot-single & 0.27 (.02) & 0.34 (.01) & 0.14 (.01) & 0.22 (.01) & 0.01 (.00) & 0.22 (.01) & 0.21 (.01) \\
Ministral-5-shot-multi & 0.45 (.01) & 0.31 (.01) & 0.22 (.01) & 0.19 (.02) & 0.26 (.01) & 0.18 (.02) & 0.16 (.01) \\
Llama-PEFT-single & \textbf{0.55 (.01)} & 0.52 (.02) & 0.29 (.01) & 0.33 (.01) & 0.35 (.02) & 0.23 (.01) & 0.28 (.02) \\
Llama-PEFT-multi & 0.54 (.01) & \textbf{0.53 (.01)} & \textbf{0.29 (.02)} & \textbf{0.41 (.02)} & \textbf{0.37 (.02)} & 0.32 (.04) & \textbf{0.26 (.05)} \\
\bottomrule
\end{tabular}
}%
\end{table*}

\begin{table*}[h!]
\centering
\footnotesize
\caption{Model Recall (mean $\pm$ sd) by Moral Category on the full MFRC dataset.}
\label{tab:model_recall}
\resizebox{.7\textwidth}{!}{%
\begin{tabular}{lccccccc}
\toprule
{Model} & {Care} & {Equality} & {Prop} & {Loyalty} & {Authority} & {Purity} & {Thin} \\
\midrule
BERT & \textbf{0.66 (.04)} & \textbf{0.56 (.05)} & \textbf{0.33 (.04)} & \textbf{0.41 (.04)} & \textbf{0.41 (.06)} & \textbf{0.42 (.09)} & \textbf{0.44 (.05)} \\
ML-BERT & 0.58 (.04) & 0.53 (.05) & 0.26 (.06) & 0.37 (.06) & 0.30 (.05) & 0.39 (.06) & 0.30 (.05) \\
Llama-0-shot-single & 0.36 (.01) & 0.27 (.01) & 0.20 (.01) & 0.16 (.01) & 0.18 (.01) & 0.16 (.01) & 0.22 (.01) \\
Llama-0-shot-multi & 0.26 (.01) & 0.30 (.01) & 0.19 (.02) & 0.13 (.01) & 0.17 (.01) & 0.12 (.02) & 0.01 (.00) \\
Ministral-0-shot-single & 0.22 (.01) & 0.24 (.01) & 0.14 (.02) & 0.20 (.01) & 0.08 (.01) & 0.09 (.01) & 0.07 (.01) \\
Ministral-0-shot-multi & 0.38 (.01) & 0.25 (.01) & 0.18 (.01) & 0.16 (.01) & 0.21 (.01) & 0.15 (.01) & 0.18 (.01) \\
Llama-5-shot-single & 0.40 (.01) & 0.20 (.01) & 0.20 (.01) & 0.13 (.01) & 0.18 (.01) & 0.11 (.01) & 0.25 (.01) \\
Llama-5-shot-multi & 0.31 (.01) & 0.31 (.01) & 0.13 (.02) & 0.16 (.01) & 0.17 (.01) & 0.13 (.02) & 0.01 (.00) \\
Ministral-5-shot-single & 0.25 (.02) & 0.31 (.01) & 0.12 (.01) & 0.20 (.01) & 0.00 (.00) & 0.19 (.01) & 0.19 (.01) \\
Ministral-5-shot-multi & 0.42 (.01) & 0.28 (.01) & 0.20 (.01) & 0.17 (.02) & 0.23 (.01) & 0.15 (.02) & 0.14 (.01) \\
Llama-PEFT-single & 0.52 (.01) & 0.49 (.02) & 0.27 (.01) & 0.31 (.01) & 0.32 (.02) & 0.20 (.01) & 0.26 (.02) \\
Llama-PEFT-multi & 0.51 (.01) & 0.50 (.01) & 0.27 (.02) & 0.39 (.02) & 0.34 (.02) & 0.30 (.04) & 0.24 (.05) \\
\bottomrule
\end{tabular}
}%
\end{table*}

%% file: TTC.tex
\section{Theory Trace Card}
See theory trace card below. 
\begin{figure*}
    
\begin{tcolorbox}[
  title={Theory Trace Card {\hypersetup{citecolor=white}\citep{karimi2026theory}} for the \textit{Moral Foundations Reddit Corpus}},
  segmentation hidden,
  enhanced,
  width=\textwidth,
  fontupper=\small
]

\textbf{1. Theory}
\begin{itemize}
  \item \textbf{Framework:} Revised Moral Foundations Theory (MFT) \citep{atari2023morality}, a pluralistic account of moral cognition identifying multiple distinct moral domains.
  \item \textbf{Core components:}
  \begin{itemize}
    \item Care/Harm.
    \item Equality/Inequality.
    \item Proportionality/Disproportionality.
    \item Loyalty/Betrayal.
    \item Authority/Subversion.
    \item Purity/Degradation.
    \item Thin Morality (pragmatic category for moral judgment not clearly grounded in a specific foundation).
  \end{itemize}
\end{itemize}

\vspace{0.5em}

\textbf{2. Components Exercised}
\begin{itemize}
  \item Detection of foundation-specific moral sentiment in text.
  \item Multi-label recognition of co-occurring moral foundations.
  \item Identification of Thin Morality.
\end{itemize}

\vspace{0.5em}

\textbf{3. Task Operationalization}
\begin{itemize}
  \item \textbf{Task:} Given a Reddit comment, the model predicts the presence or absence of each moral foundation and Thin Morality.
  \item \textbf{Key specs:} English-language Reddit comments sampled from political and everyday discourse. Each comment labeled by $\geq$3 trained annotators using a revised MFT coding manual. Majority vote aggregation determines final labels. Multi-label classification setting.
  \item \textbf{Scoring Criterion:} Standard classification metrics (e.g., F1, precision, recall) computed against majority human annotations.
\end{itemize}

\vspace{0.5em}

\textbf{4. Inference and Limitations}
\begin{itemize}
  \item \textbf{Inference:} Performance supports moral sentiment classification in online discourse under revised MFT.
  \item \textbf{Limitations:} Does not evaluate moral reasoning, normative correctness, or cross-theoretical moral competence. Operationalizes morality strictly within revised MFT; competing moral theories are outside scope. English-language Reddit data and annotator demographics may limit cultural generalizability.
\end{itemize}

\end{tcolorbox}
\end{figure*}